\documentclass{article}
\usepackage{ijcai20}

\usepackage{times}

\usepackage{soul}
\usepackage{url}
\usepackage[hidelinks]{hyperref}
\usepackage[utf8]{inputenc}
\usepackage[small]{caption}
\usepackage{graphicx}
\usepackage{amsmath}
\usepackage{booktabs}

\usepackage{algpseudocode}
\usepackage{appendix}
\usepackage{amsthm}

\usepackage{subcaption}
\usepackage{todonotes}
\usepackage{blindtext} 
\usepackage{algorithm}
\usepackage{amsfonts}   
\usepackage{mleftright}

\algnewcommand\RETURN{\State \algorithmicreturn}

\DeclareMathOperator{\E}{\mathbb{E}} 
\def\Pr{\mathbf{Pr}}
\DeclareMathOperator*{\argmax}{arg\,max}

\newtheorem{definition}{Definition}

\newcommand{\alg}{Power-UCT}

\newtheorem{innercustomgeneric}{\customgenericname}
\providecommand{\customgenericname}{}
\newcommand{\newcustomtheorem}[2]{%
  \newenvironment{#1}[1]
  {%
   \renewcommand\customgenericname{#2}%
   \renewcommand\theinnercustomgeneric{##1}%
   \innercustomgeneric
  }
  {\endinnercustomgeneric}
}

\newcustomtheorem{manualtheorem}{Theorem}
\newcustomtheorem{manuallemma}{Lemma}
\newcustomtheorem{manualassumption}{Assumption}

\newcommand{\citet}[1]{\citeauthor{#1} \shortcite{#1}}
\newcommand{\citep}{\cite}

\title{Generalized Mean Estimation in Monte-Carlo Tree Search\footnote{Copyright International Joint Conferences on Artificial Intelligence (IJCAI). All rights reserved.}}

\author{
Tuan Dam$^1$\and
Pascal Klink$^1$\and
Carlo D'Eramo$^1$\and
Jan Peters$^{1,2}$\And
Joni Pajarinen$^{1,3}$\\
\affiliations
$^1$Department of Computer Science, Technische Universit{\"a}t Darmstadt, Germany\\
$^2$Robot Learning Group, Max Planck Institute for Intelligent Systems,T{\"u}bingen, Germany\\
$^3$Computing Sciences, Tampere University, Finland\\
\emails
\{dam, klink, deramo, peters\}@ias.tu-darmstadt.de,
joni.pajarinen@tuni.fi 
}

\begin{document}

\maketitle

\begin{abstract}
We consider Monte-Carlo Tree Search (MCTS) applied to Markov Decision
Processes (MDPs) and Partially Observable MDPs (POMDPs), and the
well-known Upper Confidence bound for Trees (UCT) algorithm. In UCT, a
tree with nodes (states) and edges (actions) is incrementally built by
the expansion of nodes, and the values of nodes are updated through a
backup strategy based on the average value of child nodes. However, it
has been shown that with enough samples the maximum operator yields
more accurate node value estimates than averaging. Instead of settling for one of these value
estimates, we go a step
further proposing a novel backup strategy which uses the power mean
operator, which computes a value between the average
and maximum value. We call our new approach \alg, and argue how the
use of the power mean operator helps to speed up the learning in
MCTS. We theoretically analyze our method providing guarantees of
convergence to the optimum. Finally, we
empirically demonstrate the effectiveness of our method in well-known
MDP and POMDP benchmarks, showing significant improvement in
performance and convergence speed w.r.t. state of the art algorithms.
\end{abstract}


\section{Introduction}
\label{introduction}
Monte-Carlo Tree Search (MCTS)~\cite{coulom2006efficient} is an effective strategy for combining Monte-Carlo search with an incremental tree structure. MCTS is becoming increasingly popular in the community, especially after the outstanding results recently achieved in the game of Go~\cite{silver2016mastering}. In the last years, the MCTS research has mainly focused on effective ways of expanding the tree, performing rollouts, and backing up the average reward computed from rollouts to the parent nodes.
We consider the Upper Confidence bound applied to Trees (UCT) algorithm~\cite{kocsis2006improved}, which combines tree search with the well-known UCB1 sampling policy~\cite{auer2002finite}, as an effective way of dealing with the action selection to expand the tree. In UCT, the estimate of the value of each node is computed performing multiple rollouts starting from the node, and updating the node's value as the average of the collected rewards; then, the node's value is backed up to the parent nodes that are updated with the average of the value of the children nodes. Since the action selection policy tends to favor the best actions in the long run, UCT has theoretical convergence guarantees to the optimal value. However, it has been shown that using the average reward for backup leads to an underestimation of the optimal value, slowing down the learning; on the other hand, using the maximum reward leads to an overestimation causing the same learning problems, especially in stochastic settings~\cite{coulom2006efficient}. This problem is also evinced in the well-known $Q$-Learning algorithm~\cite{watkins:phd}, where the maximum operator leads to overestimation of the optimal value~\cite{smith2006optimizer}.
Some variants of $Q$-Learning based on (weighted) mean operators have been successfully proposed to address this issue~\cite{hasselt2010double,d2016estimating}.

In this paper, we introduce a novel backup operator based on a power mean~\cite{bullen2013handbook} that, through the tuning of a single coefficient, computes a value between the average reward and the maximum one. This allows to balance between the negatively biased estimate of the average reward, and the positively biased estimate of the maximum reward; in practice, this translates in balancing between a safe but slow update, and a greedy but misleading one. In the following, we propose a variant of UCT based on the power mean operator, which we call \alg. We theoretically prove the convergence of \alg, based on the consideration that the algorithm converges for all values between the range computed by the power mean. 
We empirically evaluate \alg~w.r.t.~UCT and the recent MENTS algorithm~\cite{xiao2019maximum} in classic MDP and POMDP benchmarks.
Remarkably, we show how \alg~outperforms the baselines both in terms of quality and speed of learning.
Thus, our \textit{contribution} is twofold:
\begin{enumerate}
    \item We propose a new backup operator for UCT based on a power mean, and prove the convergence to the optimal values;
    \item We empirically evaluate the effectiveness of our approach comparing it with UCT in well-known MDPs and POMDPs, showing significantly better performance.
\end{enumerate}
The rest of this paper is organized as follows. First we describe related work. Next, we discuss background knowledge of MCTS, UCB and UCT. Then, we describe the power mean operator and introduce our \alg~algorithm. We derive theoretical results and prove convergence to the optimum for \alg. Finally, we present empirical results in both MDP and POMDP problems, showing that \alg~outperforms baselines in MCTS.
\section{Related Work}
\label{relatedwork}







Several works focus on adapting how UCB1~\cite{auer2002finite} is applied to MCTS. For this purpose
UCB1-tuned \cite{auer2002finite} modifies the upper confidence bound of UCB1 to account for variance in order to improve exploration. 
\citet{tesauro2012bayesian} propose a Bayesian version of UCT, which obtains better estimates of node values and uncertainties given limited experience. However,
the Bayesian version of UCT is more computation-intensive.
While most work on bandits in MCTS focuses on discrete actions,
work on continuous action MCTS also exists~\cite{mansley2011sample}.
Since our MCTS algorithm is based on the UCT
algorithm, which is an extension of UCB1, our
method could be applied to all of these MCTS algorithms.







Many heuristic approaches based on specific domain knowledge have been proposed, such as adding a bonus term to value estimates based on domain
knowledge~\cite{gelly2006exploration,teytaud2010huge,childs2008transpositions,kozelek2009methods,chaslot2008progressive}
or prior knowledge collected during policy
search~\cite{gelly2007combining,helmbold2009all,lorentz2010improving,tom2010investigating,hoock2010intelligent}.
We point out that we provide a novel node value backup approach that
could be applied in combination with all of these methods.


To improve upon UCT algorithm in MCTS,
\citet{khandelwal2016analysis} formalizes and analyzes different on-policy and off-policy complex backup approaches for MCTS planning based on techniques in the Reinforcement Learning literature. \citet{khandelwal2016analysis} propose four complex backup strategies: {MCTS}$(\lambda)$,
{MaxMCTS}$(\lambda)$, {MCTS}$_\gamma$, {MaxMCTS}$_\gamma$. \citet{khandelwal2016analysis} report that {MaxMCTS}$(\lambda)$ and {MaxMCTS}$_\gamma$ perform better than UCT for certain setup of parameter. \citet{vodopivec2017monte} proposed an approach called
SARSA-UCT, which performs the dynamic programming backups using SARSA~\cite{rummery1995problem}.
Both \citet{khandelwal2016analysis} and \citet{vodopivec2017monte} directly
borrow value backup ideas from Reinforcement Learning in order to estimate
the value at each tree node.
However, they do not provide any proof of convergence.

Instead, our method provides a completely novel way of backing up values
in each MCTS node using a power mean operator, for which we prove the convergence to
the optimal policy in the limit. 
The recently introduced MENTS algorithm~\cite{xiao2019maximum}, uses softmax backup operator at each node in combination with E2W policy, and shows better convergence rate w.r.t. UCT. Given its similarity to our approach, we empirically compare to it in the experimental section.


\section{Background}\label{S:background}
In this section, we first discuss an overview of Monte Carlo Tree Search method. Next, we discuss UCB algorithm and subsequently an extension of UCB
to UCT algorithm. Finally, we discuss the definition of Power Mean operator and its properties.
\subsection{Monte-Carlo Tree Search}
MCTS combines tree search with Monte-Carlo sampling in order to build a tree, where states and actions are respectively modeled as nodes and edges, to compute optimal decisions. MCTS requires a generative black box simulator for generating a new state based on the current state and chosen action. The MCTS algorithm consists of a loop of four steps:
\begin{itemize}
\item[--] \textbf{Selection:} start from the root node, interleave action selection and sampling the next state (tree node) until a leaf node is reached
\item[--] \textbf{Expansion:} expand the tree by adding a new edge (action) to the leaf node and sample a next state (new leaf node)
\item[--] \textbf{Simulation:} rollout from the reached state to the end of the episode using random actions or a heuristic
\item[--] \textbf{Backup:} update the nodes backwards along the trajectory starting from the end of the episode until the root node according to the rewards collected
\end{itemize}
In the next subsection, we discuss UCB algorithm and its extension to UCT.



\subsection{Upper Confidence Bound for Trees}

In this work, we consider the MCTS algorithm UCT (Upper Confidence bounds for
Trees)~\cite{kocsis2006improved}, an extension
of the well-known UCB1~\cite{auer2002finite} multi-armed bandit algorithm. UCB1 chooses
the arm (action $a$) using
\begin{flalign}
a = \argmax_{i \in \{1...K\}} \overline{X}_{i, T_i(n-1)} + C\sqrt{\frac{\log n}{T_i(n-1)}}.
\label{UCB1}
\end{flalign}
where $T_i(n) = \sum^n_{t=1} \textbf{1} \{t=i\} $ is the number of
times arm $i$ is played up to time $n$. $\overline{X}_{i, T_i(n-1)}$
denotes the average reward of arm $i$ up to time $n-1$ and $C
= \sqrt{2}$ is an exploration constant. 
In UCT, each node is a separate bandit, where the arms correspond to the actions, and the payoff is the reward of the episodes starting from them.
In the backup phase, value is backed up recursively from the leaf node to the root as
\begin{flalign}
\overline{X}_n = \sum^{K}_{i=1} \Big(\frac{T_i(n)}{n}\Big) \overline{X}_{i, T_i(n)}.
\end{flalign}
\citet{kocsis2006improved} proved that UCT converges in the limit to the optimal policy.

\subsection{Power Mean}
In this paper, we introduce a novel way of estimating the
expected value of a bandit arm ($\overline{X}_{i, T_i(n-1)}$ in
(\ref{UCB1})) in MCTS. For this purpose, we will use the \textit{power
mean}~\cite{mitrinovic1970analytic}, an operator belonging to the family of functions for aggregating sets of numbers, that includes as special cases the Pythagorean means (arithmetic, geometric, and harmonic means):
\begin{definition}
For a sequence of positive numbers $X = (X_1,...,X_n)$ and positive weights $w = (w_1,...,w_n)$, the power 
mean of order $p$ ($p$ is an extended real number) is defined as
\begin{flalign}
\text{M}^{[p]}_n(X,w) = \Bigg( \frac{\sum^n_{i=1} w_i X_i^p}{\sum^n_{i=1} w_i} \Bigg)^{\frac{1}{p}}.
\label{E:power_mean}
\end{flalign}
\end{definition}
\noindent With $p = 1$ we obtain the weighted arithmetic mean. With $p \rightarrow 0$ we have the
geometric mean, and with $p = -1$ we have the harmonic mean \cite{mitrinovic1970analytic}
Furthermore, we get~\cite{mitrinovic1970analytic}
\begin{flalign}
\text{M}^{[-\infty]}_n(X,w) = \lim_{p\rightarrow -\infty}  \text{M}^{[p]}_n(X,w) = \text{Min}(X_1,...,X_n), \\
\text{M}^{[+\infty]}_n(X,w) = \lim_{p\rightarrow +\infty}  \text{M}^{[p]}_n(X,w) = \text{Max}(X_1,...,X_n),
\end{flalign}

The weighted arithmetic mean lies between
$\text{Min}(X_1,...,X_n)$ and $\text{Max}(X_1,...,X_n)$. Moreover,
the following lemma shows that $\text{M}^{[p]}_n(X,w)$ is an increasing function.
\begin{manuallemma}{1}
$\text{M}^{[p]}_n(X,w)$ is an increasing function meaning that
\begin{flalign}
\text{M}^{[1]}_n(X,w) \leq \text{M}^{[q]}_n(X,w) \leq \text{M}^{[p]}_n(X,w),\forall p \geq q \geq 1
\end{flalign}
\end{manuallemma}
\begin{proof}
\noindent For the proof, see~\cite{mitrinovic1970analytic}.
\end{proof}
The following lemma shows that Power Mean can be upper bound by Average Mean plus with a constant.
\begin{manuallemma}{2}\label{lb:lemma2}
Let $0 < l \leq X_i \leq U, C = \frac{U}{l}, \forall i \in (1, ..., n) $ and $p > q$. We define:
\begin{flalign}
\text{Q}(X, w, p, q) &= \frac{\text{M}^{[p]}_n(X,w)}{\text{M}^{[q]}_n(X,w)}\\
\text{D}(X, w, p, q) &= \text{M}^{[p]}_n(X,w) - \text{M}^{[q]}_n(X,w).
\end{flalign}
Then we have:
\begin{flalign}
&\text{Q}(X, w, p, q) \leq \text{L}_{p,q} \text{D}(X, w, p, q) \leq \text{H}_{p,q} \nonumber \\
&\text{L}_{p,q} = \Bigg( \frac{q(C^p - C^q)}{(p-q)(C^q - 1)}\bigg)^{\frac{1}{p}} 
\Bigg( \frac{p(C^q - C^p)}{(q-p)(C^p - 1)} \bigg)^{-\frac{1}{q}} \nonumber \\
&\text{H}_{p,q} = (\theta U^p + (1 - \theta) l^p)^{\frac{1}{p}} - (\theta U^q + (1 - \theta) l^q)^{1/q}, \nonumber
\end{flalign}

\noindent where $\theta$ is defined in the following way. Let
\begin{flalign}
h(x) &= x^{\frac{1}{p}} - (ax + b)^{1/q} \nonumber
\end{flalign}
where:
\begin{flalign}
 &a = \frac{U^q - l^q}{U^p - l^p}; b = \frac{U^p l^q - U^q l^p}{U^p - l^p}\\
 &x^{'} = \argmax \{h(x), x \in (l^p, U^p)\}
\end{flalign}
then:
\begin{flalign}
\theta = \frac{x' - l^p}{U^p - l^p}.\nonumber
\end{flalign}
\end{manuallemma}
\begin{proof}
Refer to \citet{mitrinovic1970analytic}.
\end{proof}{}



\section{Power Mean Backup}
As previously described, it is well known that performing backups using the average of the rewards results in an underestimate of the true value of the node, while using the maximum results in an overestimate of it~\cite{coulom2006efficient}. Usually, the average backup is used when the number of simulations is low, for a conservative update of the nodes due to the lack of samples; on the other hand, the maximum operator is favoured when the number of simulations is high. We address this problem proposing a novel backup operator for UCT based on the power mean (Equation~\ref{E:power_mean}):
\begin{flalign}
\overline{X}_n(p) = \left(\sum^{K}_{i=1} \left(\frac{T_i(n)}{n}\right) \overline{X}_{i, T_i(n)}^p\right)^{\frac{1}{p}}.
\end{flalign}
This way, we bridge the gap between the average and maximum estimators with the purpose of getting the advantages of both. We call our approach \alg~and describe it in more detail in the following.
\subsection{\alg}
\begin{algorithm}
\caption{\alg}
\label{power_uct}
\begin{algorithmic}[1]
\State {\textbf{Input}}
\State {$s$: {state}}
\State {$a$: {action}}
\State {$p$: {Power Mean constant}}
\State {$C$: {Exploration constant}}
\State {$N(s)$: {number of simulations of V\_Node of state $s$}}
\State {$n(s,a)$: {number of simulations of Q\_Node of state $s$ and action $a$}}
\State {$V(s)$: {Value of V\_Node at state $s$. Default is 0}}
\State {$Q(s,a)$: {Value of Q\_Node at state $s$, action $a$. Default is 0}}
\State {$\tau(s,a)$: {transition function}}
\State {$\gamma$: {discount factor}}
\\
\\
\Procedure{SELECT\_ACTION}{s}
\RETURN{} $\argmax_{a} Q(s, a) + C\sqrt{\frac{\log N(s)}{n(s,a)}}$
\EndProcedure
\\
\Procedure{SEARCH}{s}
\Repeat
\State{SIMULATE\_V(s, 0)}
\Until{TIMEOUT()}
\RETURN{} $\argmax_a  Q(s,a)$
\EndProcedure
\\
\Procedure{Rollout}{s, depth}
\If {$ \gamma^{\text{depth}} < \epsilon$}
\RETURN{} $0$
\EndIf
\State {$a \sim \pi_{\text{rollout}}(.)$}
\State {$(s', r) \sim $}{ $\tau(s,a)$}
\RETURN{} $r + \gamma${ROLLOUT} $(s',depth + 1)$
\EndProcedure
\\
\Procedure{SIMULATE\_V}{s, depth}
\State {$a \gets ${SELECT\_ACTION}$(s)$}
\State {{SIMULATE\_Q} $(s,a, \text{depth})$}
\State {$N(s) \gets N(s) + 1$}
\State {$V(s) \gets \left(\sum_{a} \frac{n(s,a)}{N(s)} Q(s,a)^p\right)^{\frac{1}{p}}$}
\EndProcedure
\\
\Procedure{SIMULATE\_Q}{s, a, depth}
\State {$(s',r)$}{$ \sim \tau(s,a)$}
\If {$ s' \notin {Terminal} $}
\If {$V(s') \text{ not expanded}$}
\State {{ROLLOUT}$(s',\text{depth})$}
\Else
\State {{SIMULATE\_V}$(s',\text{depth} + 1)$}
\EndIf
\EndIf
\State {$n(s,a) \gets n(s,a) + 1$}
\State {$Q(s,a) \gets \frac{(\sum r_{s,a}) + \gamma.\sum_{s'} N(s').V(s')}{n(s,a)}$}
\EndProcedure
\end{algorithmic}
\end{algorithm}
The introduction of our novel backup operator in UCT does not require major changes to the algorithm. Indeed, the \alg~pseudocode shown in Algorithm~\ref{power_uct} is almost identical to the UCT one, with the only differences in lines 36 and 49. MCTS has two type
of nodes: V\_Nodes corresponding to the state-value, and Q\_Nodes corresponding to state-action values. An action is taken from the V\_Node of the current state leading to the respective Q\_Node, then it leads to the V\_Node of the reached state. We skip the description of all the procedures since they are well-known components of MCTS, and we focus only on the ones involved in the use of the power mean backup operator.
In SIMULATE\_V, \alg~updates the value of each V\_Node using the power mean of its children Q\_Nodes, that are computed in SIMULATE\_Q. Note that our algorithm could be applied to several bandit based enhancements of UCT, but for simplicity we only focus on UCT.
\section{Theoretical Analysis}
In this section, we show that \alg~ can smoothly adapt to all theorems of UCT~\cite{kocsis2006improved}. The following results can be seen as a generalization of the results for UCT, as we consider a generalized mean instead of a standard mean as the backup operator.
Our main results are Theorem 6 and Theorem 7, which respectively prove the convergence of failure probability at the root node, and derive the bias of power mean estimated payoff. 
In order to prove them, we start with Theorem \ref{th:th_5} to show the concentration of power mean with respect to i.i.d random variables $X$. Subsequently, Theorem 2 shows the upper bound of the expected number of times when a suboptimal arm is played.
Theorem 3 bounds the expected error of the power mean estimation.
Theorem 4 shows the lower bound of the number of times any arm is played.
Theorem 5 shows the concentration of power mean backup around its mean value at each node in the tree\footnote{Refer to \cite{dam2019generalized} for proofs of lemmas and theorems.}.

\begin{manualtheorem}{1} \label{th:th_5}
If $X_1, X_2, ..., X_n$ are independent with $\Pr(a \leq X_i \leq b) = 1$
and common mean $\mu$, $w_1, w_2, ..., w_n$ are positive and $W = \sum^n_{i=1} w_i$ then for any $\epsilon > 0$, $p \geq 1$
\begin{flalign}
\Pr \Bigg ( \Big| \Big( \frac{\sum^n_{i=1} w_i X_i^p}{\sum^n_{i=1} w_i} \Big)^{\frac{1}{p}} - \mu \Big | > \epsilon \bigg) \nonumber \\
\leq 2 \exp(\text{H}_{p, 1})\exp(-2\epsilon^2 W^2 / \sum^n_{i=1} w_i^2 (b-a)^2)) \nonumber
\end{flalign}
\end{manualtheorem}

\noindent Theorem~\ref{th:th_5} is derived using the upper bound of the power mean operator, which corresponds to the average mean incremented by a constant~\cite{mitrinovic1970analytic} and Chernoff's inequality.
\noindent Note that this result can be considered a generalization of the well-known Hoeffding inequality to power mean. Next, given i.i.d.\ random variables $X_{it}$ (t=1,2,...) as the payoff sequence at any internal leaf node of the tree, we assume the expectation of the payoff exists and let $\mu_{in} = \E[\overline{X_{in}}]$.  We assume the power mean reward drifts as a function of time and converges only in the limit, which means that
\begin{flalign}
\mu_{i} = \lim_{n \rightarrow \infty} {\mu_{in}}. \nonumber
\end{flalign}
\noindent Let $\delta_{in} = \mu_{i} - \mu_{in}$ which also means that 
\begin{flalign}
\lim_{n \rightarrow \infty} {\delta_{in}} = 0. \nonumber
\end{flalign}
From now on, let $*$ be the upper index for all quantities related to the optimal arm. By assumption, the rewards lie between $0$ and $1$. Let's start with an assumption:
\begin{manualassumption}{1}\label{asumpt}
Fix $1 \leq i \leq K$. Let $\{F_{it}\}_t$ be a filtration such that $ \{X_{it}\}_t$ is $\{F_{it}\}$-adapted
and $X_{i,t}$ is conditionally independent of $F_{i,t+1}, F_{i,t+2},...$ given $F_{i,t-1}$. Then $0 \leq X_{it} \leq 1$ and the limit of $\mu_{in} = \E[\overline{X_{in}}(p)]$ exists. Further, we assume that there exists a constant $C > 0$ and an integer $N_c$ such that for $n>N_c$, for any $\delta > 0$, $ \triangle_n(\delta) = C\sqrt{n\log(1/\delta)}$, the following bounds hold:
\begin{flalign}
\Pr(\overline{X}_{in}(p) \geq \E[ \overline{X}_{in}(p)] + \triangle_n(\delta)/n) \leq \delta \label{eq:3}, \\
\Pr(\overline{X}_{in}(p) \leq \E[ \overline{X}_{in}(p)] - \triangle_n(\delta)/n) \leq \delta \label{eq:4}.
\end{flalign}
\end{manualassumption}
\noindent Under Assumption 1, a suitable choice for the bias sequence $c_{t,s}$ is given by 
\begin{flalign}
c_{t,s} = 2C\sqrt{\frac{\log{t}}{s}}. \label{eq:8}
\end{flalign}
where C is an exploration constant.\\
\noindent Next, we derive Theorems~\ref{T:th_2}, \ref{T:th_3}, and~\ref{T:th_4} following the derivations in~\cite{kocsis2006improved}.
First, from Assumption \ref{asumpt}, we derive an upper bound on the error for the expected number of times suboptimal arms are played.
\begin{manualtheorem} {2}\label{T:th_2}
Consider UCB1 (using power mean estimator) applied to a non-stationary problem where the pay-off sequence satisfies Assumption 1 and 
where the bias sequence, $c_{t,s}$ defined in (\ref{eq:8}). Fix $\epsilon \geq 0$. Let $T_k(n)$ denote the number of plays of arm $k$. Then if $k$ is the index of a suboptimal arm then each sub-optimal arm $k$ is played in expectation at most
\begin{flalign}
\E[T_k(n)] \leq \frac{16C^2\ln n}{(1-\epsilon)^2 \triangle_k^2} + A(\epsilon) + N_c + \frac{\pi^2}{3} + 1.
\end{flalign}
\end{manualtheorem}
\noindent Next, we derive our version of Theorem 3 in~\cite{kocsis2006improved}, which computes the upper bound of the difference between the value backup of an arm with $\mu^*$ up to time $n$.
\begin{manualtheorem} {3}\label{T:th_3}
Under the assumptions of Theorem~\ref{T:th_2},
\begin{flalign}
\big| \E\big[ \overline{X}_n(p) \big]  - \mu^{*} \big| &\leq |\delta^*_n| + \mathcal{O} \Bigg( \frac{K(C^2 \log n + N_0)}{n} \Bigg)^{\frac{1}{p}}. \nonumber
\end{flalign}
\end{manualtheorem}
\noindent A lower bound for the times choosing any arm follows:
\begin{manualtheorem} {4}\label{T:th_4} (\textbf{Lower Bound})
Under the assumptions of Theorem 2, there exists some positive constant $\rho$ such that for all arms k and n,
$T_k(n) \geq \lceil \rho \log (n)\rceil$.
\end{manualtheorem}
\noindent For deriving the concentration of estimated payoff around its mean,
we modify Lemma 14 in~\cite{kocsis2006improved} for power mean: in the proof, we 
first replace the partial sums term with a partial mean term
and modify the following equations accordingly. The partial mean term can then be easily replaced by a partial power mean term and we get
\begin{manuallemma}{7} \label{lb:lemma7}
Let $Z_i$, $a_i$ be as in Lemma 13 in~\cite{kocsis2006improved}.
Let $F_i$ denotes a filtration over some probability space. $Y_i$ be an $F_i$-adapted real valued martingale-difference sequence. Let {$X_i$} be an i.i.d.\ sequence with mean 
$\mu$. We assume that both $X_i$ and $Y_i$ lie
in the [0,1] interval. Consider the partial sums\\
\begin{flalign}
S_n = \Bigg(\frac{\sum_{i=1}^n (1-Z_i) X_i^p + Z_i Y_i^p}{n}\Bigg)^{\frac{1}{p}}.
\end{flalign}
Fix an arbitrary $\delta > 0$, and fix $p \geq 1$, and $M = \exp(H_{p,1})$ where $H_{p,1}$ is defined as in Lemma \ref{lb:lemma2}. Let $\triangle_n = 9\sqrt{2n \log(2M/\delta)}$, and $\triangle = (9/4)^{p-1}\triangle_{n}$ let 
\begin{flalign}
R_n = \E\Bigg[\Bigg(\frac{\sum_{i = 1}^n X_i^p}{n}\Bigg)^{\frac{1}{p}}\Bigg] - \E[S_n] \label{23}.
\end{flalign}
Then for n such that $a_n \leq (1/9)\triangle_n$ and $|R_n| \leq (4/9) (\triangle/n)^{\frac{1}{p}}$
\begin{flalign}
\Pr(S_n \geq \E[S_n] + (\triangle/n)^{\frac{1}{p}}) \leq \delta \label{lb_lower}\\
\Pr(S_n \leq \E[S_n] - (\triangle/n)^{\frac{1}{p}}) \leq \delta \label{lb_upper}
\end{flalign}
\end{manuallemma}
Based on the results from Lemma \ref{lb:lemma7}, we derive the concentration of estimated payoff around its mean.
\begin{manualtheorem} {5} \label{theorem5}
Fix an arbitrary $\delta > 0$ and fix $p \geq 1$, $M = \exp(H_{p,1})$ where $H_{p,1}$ is defined as in Lemma \ref{lb:lemma2} and let $\triangle_n = (\frac{9}{4})^{p-1} (9\sqrt{2n \log(2M/\delta)})$. Let $n_0$ be such that
\begin{flalign}
\sqrt{n_0} \leq \mathcal{O}(K(C^2 \log n_0 + N_0 (1/2))).
\end{flalign}
Then for any $n \geq n_0$, under the assumptions of Theorem 2, the following bounds hold true:
\begin{flalign}
\Pr(\overline{X}_{n}(p) \geq \E[ \overline{X}_{n}(p)] + (\triangle_n/n)^{\frac{1}{p}}) \leq \delta \\
\Pr(\overline{X}_{n}(p) \leq \E[ \overline{X}_{n}(p)] - (\triangle_n/n)^{\frac{1}{p}}) \leq \delta
\end{flalign}
\end{manualtheorem}
\noindent Using The Hoeffding-Azuma inequality for Stopped Martingales Inequality (Lemma 10 in \citet{kocsis2006improved}), under Assumption 1 and the result from Theorem 4 we get
\begin{manualtheorem} {6} (\textbf{Convergence of Failure Probability})
Under the assumptions of Theorem 2, it holds that
\begin{flalign}
\lim_{t\rightarrow \infty} \Pr(I_t \neq i^*) = 0.
\end{flalign}
\end{manualtheorem}
\noindent And finally, the following is our main result showing the expected payoff of our \alg.
\begin{manualtheorem} {7}\label{T:th_7}
Consider algorithm \alg \space running on a game tree of depth D, branching factor K with stochastic payoff
at the leaves. Assume that the payoffs lie in the interval [0,1]. Then the bias of the estimated expected
payoff, $\overline{X_n}$, is $\mathcal{O} (KD (\log (n)/n)^{\frac{1}{p}} + K^D (1/n)^{\frac{1}{p}})$. 
Further, the failure probability at the root convergences to zero as the number of samples grows to infinity.
\end{manualtheorem}
\begin{proof} (Sketch)
As for UCT~\cite{kocsis2006improved}, the proof is done by induction on $D$. When $D = 1$, \alg~corresponds to UCB1 with power mean backup, and our assumptions on the payoffs hold, thanks to Theorem~\ref{th:th_5}, the proof of convergence follows the results as Theorem 3 and Theorem 6.

Now we assume that the result holds up to depth $D-1$ and consider the tree of depth $D$.
Running \alg~on root node is equivalent to UCB1 on non-stationary bandit settings, but with power mean backup. Theorem 3 shows that the expected average payoff converges. The conditions on the exponential concentration of the payoffs (Assumption 1) are satisfied follows from Theorem 5.
The error bound of running \alg~for the whole tree is the sum of payoff at root node with payoff starting from any node $i$ after the first action chosen from root node until the end. This payoff by induction at depth $D-1$ in addition to the bound from Theorem 3 when the the drift-conditions are satisfied, and with straightforward algebra, we can compute the payoff at the depth $D$, in combination with Theorem 6. Since by our induction hypothesis this holds for all nodes at a distance of one node from the root, the proof is finished by observing that Theorem 3 and Theorem 5 do indeed ensure that the drift conditions are satisfied.
This completes our proof of the convergence of \alg. Interestingly, the proof guarantees the convergence for any finite value of $p$.
\end{proof}

\section{Experiments}

In this section, we aim to answer the following questions empirically: Does the Power Mean offer higher performance in MDP and POMDP MCTS tasks than the regular Mean? How does the value of $p$ influence the overall performance? How does Power-UCT, our MCTS algorithm based on the Power Mean, compare to state-of-the-art methods in tree-search? We choose the recent MENTS algorithm~\cite{xiao2019maximum}, that introduces a maximum entropy policy optimization framework for MCTS, and shows better convergence rate w.r.t. UCT.

For MENTS we find the best combination of the two hyper-parameters by grid search. In MDP tasks, we find the UCT exploration constant using grid search. For \alg, we find the $p$-value by increasing it until performance starts to decrease.

\subsection{\textit{FrozenLake}}


\begin{table}
\centering

\resizebox{.95\columnwidth}{!}{
\smallskip
\begin{tabular}{|l|c|c|c|c|}\hline
Algorithm & $4096$ & $16384$ & $65536$ & $262144$\\\cline{1-5}
UCT & $0.08 \pm 0.02$ & $0.23 \pm 0.04$ & $0.54 \pm 0.05$ & $0.69 \pm 0.04$\\\cline{1-5}
p=$2.2$ & $0.12 \pm 0.03$ & $0.32 \pm 0.04$ & $\mathbf{0.62 \pm 0.04}$ & $\mathbf{0.81 \pm 0.03}$\\\cline{1-5}
p=$\text{max}$ & $0.10 \pm 0.03$ & $0.36 \pm 0.04$ & $0.55 \pm 0.04$ & $0.69 \pm 0.04$\\\cline{1-5}
MENTS & $\mathbf{0.28 \pm 0.04}$ & $\mathbf{0.46 \pm 0.04}$ & $\mathbf{0.62 \pm 0.04}$ & $0.74 \pm 0.04$\\\cline{1-5}
\end{tabular} 
}
\caption{Mean and two times standard deviation of the success rate, over $500$ evaluation runs, of UCT, \alg~and MENTS in \textit{FrozenLake} from OpenAI Gym.
The top row of each table shows the number of simulations used for tree-search at each time step.}
\label{tab:frozen-lake}
\end{table}





For MDPs, we consider the well-known \textit{FrozenLake} problem as implemented in OpenAI Gym~\cite{brockman2016openai}. In this problem, the agent needs to reach a goal position in an 8x8 ice grid-world while avoiding falling into the water by stepping onto unstable spots. The challenge of this task arises from the high-level of stochasticity, which makes the agent only move towards the intended direction one-third of the time, and into one of the two tangential directions the rest of it. Reaching the goal position yields a reward of $1$, while all other outcomes (reaching the time limit or falling into the water) yield a reward of zero. As can be seen in Table~\ref{tab:frozen-lake}, \alg~improves the performance compared to UCT. Power-UCT outperforms MENTS when the number of simulations increases.

\begin{figure}
\centering
\includegraphics[scale=.65]{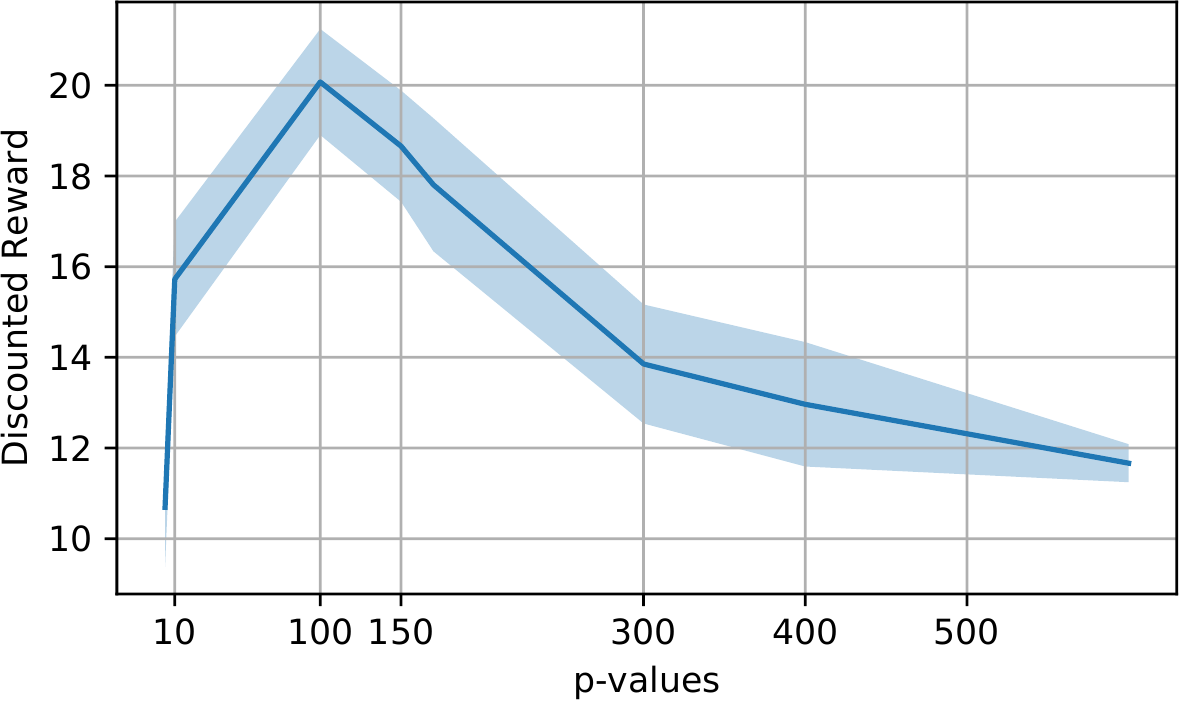}
\caption{Evaluating \alg \space w.r.t.\ different $p$-values: The mean discounted total reward at 65536 simulations (shaded area denotes standard error) over $100$ evaluation runs.}
\label{p_analysis}
\end{figure}




\begin{figure*}
\centering
\includegraphics[scale=.39]{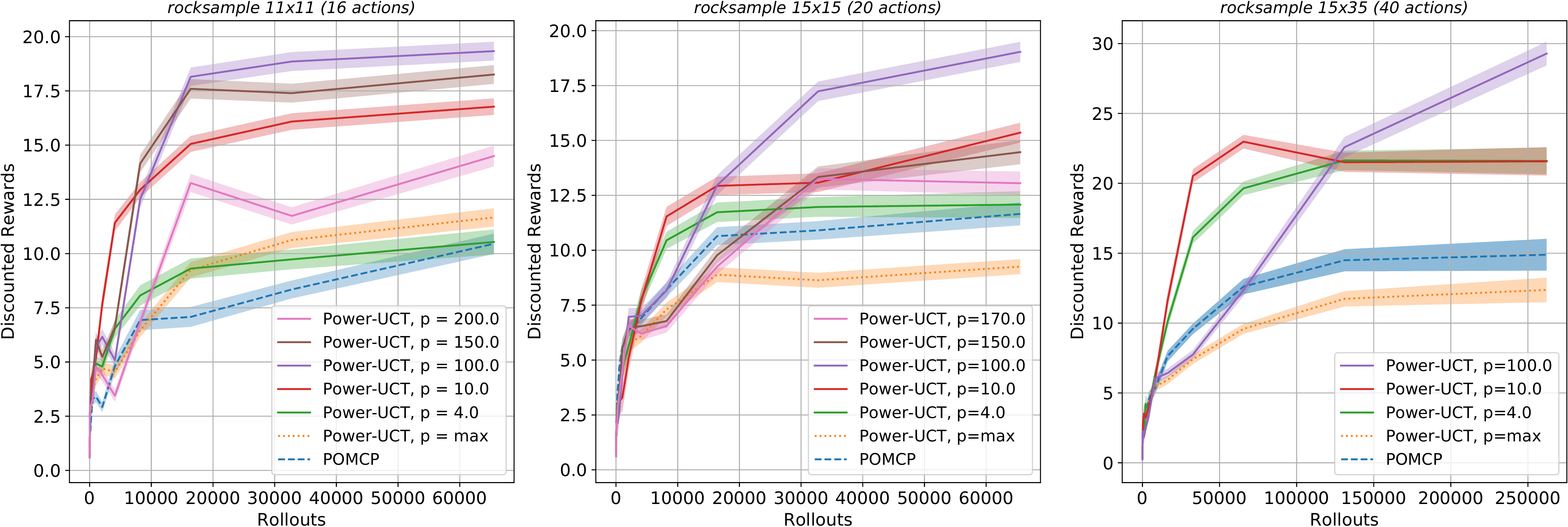}
\caption{
Performance of Power-UCT compared to UCT in \textit{rocksample}.
The mean of total discounted reward over $1000$ evaluation runs is shown by thick lines while the shaded area shows standard error.}
\label{pomcp_fig_all}
\end{figure*}

\subsection{Copy Environment}

Now, we aim to answer the question of how Power-UCT scales to domains with a large number of actions (high branching factor). We use the OpenAI gym Copy environment where the agent needs to copy the characters on an input band to an output band. The agent can move and read the input band at every time-step and decide to write a character from an alphabet to the output band. Hence, the number of actions scales with the size of the alphabet. 

Contrary to the previous experiments, there is only one initial run of tree-search and afterwards, no re-planning between two actions occurs. Hence, all actions are selected according to the value estimates from the initial search. The results in Tables~\ref{tab:copy-144} and \ref{tab:copy-200} show that Power-UCT allows to solve the task much quicker than regular UCT. Furthermore, we observe that MENTS and Power-UCT for $p=\infty$ exhibit larger variance compared to Power-UCT with a finite value of $p$ and are not able to reliably solve the task, as they do not reach the maximum reward of $40$ with $0$ standard deviation.





\begin{table}
\begin{subtable}{.48\textwidth}
\centering

\resizebox{.95\columnwidth}{!}{
\smallskip
\begin{tabular}{|l|c|c|c|c|}\hline
Algorithm & $512$ & $2048$ & $8192$ & $32768$ \\\cline{1-5}
UCT & $2.6 \pm 0.98$ & $9. \pm 1.17$ & $34.66 \pm 1.68$ & $\mathbf{40. \pm 0.}$\\\cline{1-5}
$p=3$ & $3.24 \pm 1.17$ & $\mathbf{12.35 \pm 1.14}$ & $\mathbf{40. \pm 0.}$ & $\mathbf{40. \pm 0.}$\\\cline{1-5}
$p=\text{max}$ & $2.56 \pm 1.48$ & $9.55 \pm 3.06$ & $37.52 \pm 5.11$ & $39.77 \pm 0.84$\\\cline{1-5}
MENTS & $\mathbf{3.26 \pm 1.32}$ & $11.96 \pm 2.94$ & $39.37 \pm 1.15$ & $39.35 \pm 0.95$\\\cline{1-5}
\end{tabular} 
}
\caption{144 Actions}
\label{tab:copy-144}
\end{subtable}

\begin{subtable}{.48\textwidth}
\centering

\resizebox{.95\columnwidth}{!}{
\smallskip
\begin{tabular}{|l|c|c|c|c|}\hline
Algorithm & $512$ & $2048$ & $8192$ & $32768$ \\\cline{1-5}
UCT & $1.98 \pm 0.63$ & $6.43 \pm 1.36$ & $24.5 \pm 1.56$ & $\mathbf{40. \pm 0.}$\\\cline{1-5}
$p=3$ & $\mathbf{2.55 \pm 0.99}$ & $\mathbf{9.11 \pm 1.41}$ & $\mathbf{36.02 \pm 1.72}$ & $\mathbf{40. \pm 0.}$\\\cline{1-5}
$p=\text{max}$ & $2.03 \pm 1.37$ & $6.99 \pm 2.51$ & $27.89 \pm 4.12$ & $39.93 \pm 0.51$\\\cline{1-5}
MENTS & $2.44 \pm 1.34$ & $8.86 \pm 2.65$ & $34.63 \pm 5.6$ & $39.42 \pm 0.99$\\\cline{1-5}
\end{tabular} 
}

\caption{200 Actions}
\label{tab:copy-200}
\end{subtable}
\caption{Mean and two times standard deviation of discounted total reward, over $100$ evaluation runs, of UCT, Power-UCT and MENTS in the copy environment with 144 actions (top) and 200 actions (bottom). Top row: number of simulations at each time step.}
\label{tab:copy-results}
\end{table}

\subsection{Rocksample and PocMan}
In POMDP problems, we compare \alg \space against the POMCP algorithm~\cite{silver2010monte} which is a standard UCT algorithm for POMDPs.
Since the state is not fully observable in POMDPs, POMCP assigns a unique action-observation history, which is a sufficient statistic for optimal decision making in POMDPs, instead of the state, to each tree node. Similarly to fully observable UCT, POMCP chooses actions using the UCB1 bandit. Therefore, we modify POMCP to use the power mean identically to how we modified fully observable UCT and get a POMDP version of \alg. We also modify POMCP similarly for the MENTS approach.
Next, we discuss the evaluation of the POMDP based \alg, MENTS, and POMCP, in the \textit{rocksample} and \textit{pocman} environments~\cite{silver2010monte}. 
In both problems, we scale the rewards into [0,1] for MCTS planning and show the plot with real reward scale.

\begin{figure}
\centering
\includegraphics[scale=.63]{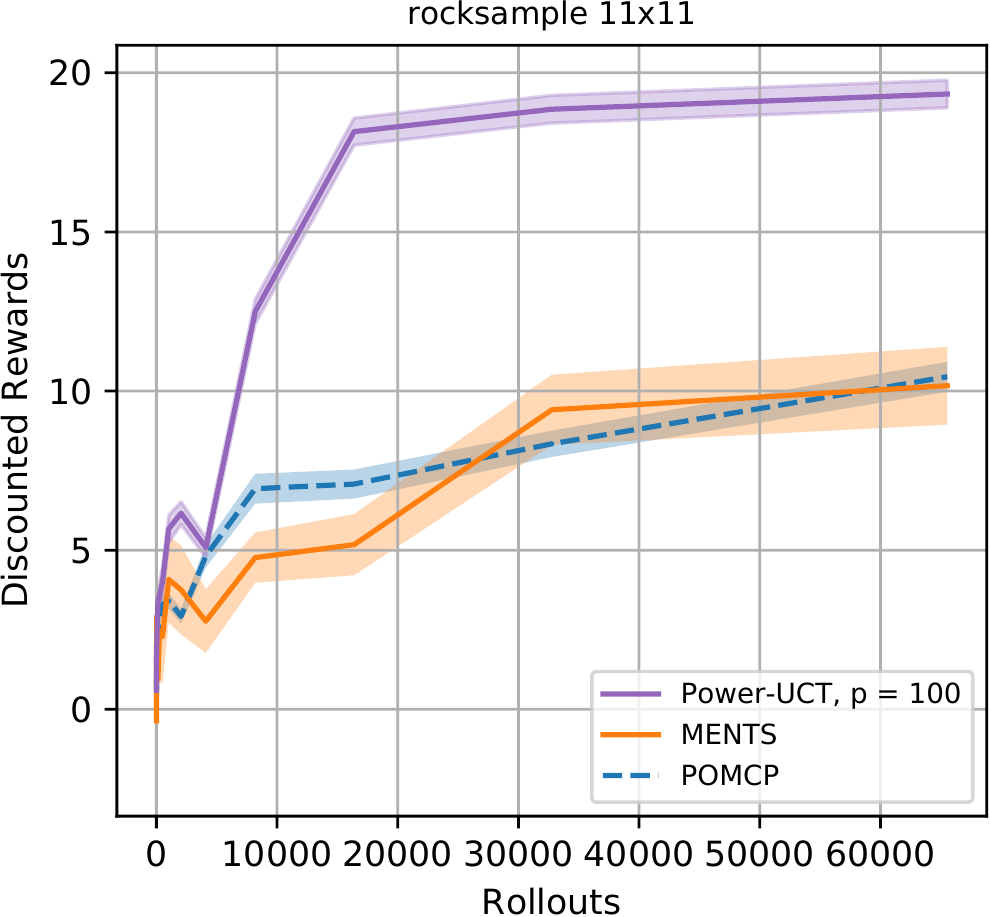}
\caption{Performance of \alg \space compared to UCT and MENTS in \textit{rocksample} 11x11. The mean of discounted total reward over $1000$ evaluation runs is shown by thick lines while the shaded area shows standard error.}
\label{rocksample11x11_fig}
\end{figure}

\paragraph{Rocksample.}
The \textit{rocksample (n,k)} (\citet{smith2004heuristic}) simulates a Mars explorer robot 
in an \textit{$n$ x $n$} grid containing \textit{k} rocks. The task is to determine
which rocks are valuable using a long range sensor, take samples of valuable rocks and to finally leave
the map to the east. There are $k + 5$ actions where the agent can move in four directions (North, South, East, West), sample a rock, or sense one of the $k$ rocks. Rocksample requires strong exploration to find informative actions which do not yield immediate reward but may yield high long term reward.
We use three variants with a different number of actions: \textit{rocksample} (11,11), \textit{rocksample} (15,15), \textit{rocksample} (15,35) and set the exploration constant as in~\cite{silver2010monte} to the difference of the maximum and minimum immediate reward.
In Fig.~\ref{pomcp_fig_all}, \alg \space outperforms POMCP for almost all values of $p$. For sensitivity analysis, Fig.~\ref{p_analysis} shows the performance of \alg \space in \textit{rocksample} (11x11) for different $p$-values at 65536 simulations. Fig.~\ref{p_analysis} suggests that at least in \textit{rocksample} finding a good $p$-value is straightforward. Fig.~\ref{rocksample11x11_fig} shows that \alg \space significantly outperforms MENTS in \textit{rocksample} (11,11). A possible explanation for the strong difference in performance between MENTS and \alg~is that MENTS may not explore sufficiently in this task. However, this would require more in depth analysis of MENTS.




\paragraph{Pocman.}
We further evaluate our algorithm in the \textit{pocman} problem~\cite{silver2010monte}. In \textit{pocman},
an agent called PocMan must travel in a maze of size (17x19) by only observing the local neighborhood in the maze. PocMan tries to eat as many food pellets as possible. Four ghosts try to kill PocMan. After moving initially randomly the ghosts start to follow directions with a high number of food pellets more likely.
If PocMan eats a power pill, he is able to eat ghosts for $15$ time steps.
PocMan receives a reward of $-1$ at each step he travels, $+10$ for eating each food pellet, $+25$ for eating a ghost and $-100$ for dying. The \textit{pocman} problem has $4$ actions, $1024$ observations, and approximately $10^{56}$ states.
Table~\ref{pocman_table} shows that \alg \space and MENTS outperform POMCP.



\begin{table}
\centering
\resizebox{.95\columnwidth}{!}{
\smallskip
\begin{tabular}{|l|c|c|c|c|}\hline
 & $1024$ & $4096$ & $16384$ & 65536 \\\cline{1-5}
$\text{POMCP}$ & $30.89 \pm 1.4$ & $33.47 \pm 1.4$ & $33.44 \pm 1.39$ & $32.36 \pm 1.6$\\\cline{1-5}
$p=max$ & $14.82 \pm 1.52$ & $14.91 \pm 1.52$ & $14.34 \pm 1.52$ & $14.98 \pm 1.76$\\\cline{1-5}
$p=10$ & $29.14 \pm 1.61$ & $35.26 \pm 1.56$ & $44.14 \pm 1.60$ & $\mathbf{53.30 \pm 1.46}$\\\cline{1-5}
$p=30$ & $28.78 \pm 1.44$ & $33.92 \pm 1.56$ & $42.45 \pm 1.54$ & $49.66 \pm 1.70$\\\cline{1-5}
$\text{MENTS}$ & $\mathbf{54.08 \pm 3.20}$ & $\mathbf{55.37 \pm 3.0}$ & $\mathbf{53.90 \pm 2.86}$ & $51.03 \pm 3.36$\\\cline{1-5}
\end{tabular}
}
\caption{Discounted total reward in \textit{pocman} for the comparison methods. Mean $\pm$ standard error are computed from $1000$ simulations except in MENTS where we ran $100$ simulations.}
\label{pocman_table}
\end{table}

\section{Conclusion}
We proposed to use power mean as a novel backup operator in MCTS, and derived a variant of UCT based on this operator, which we call \alg. We theoretically prove the convergence of \alg~to the optimal value, given that the value computed by the power mean lies between the average and the maximum. The empirical evaluation on stochastic MDPs and POMDPs, shows the advantages of \alg~w.r.t. other baselines.

Possible future work includes the proposal of a theoretically justified or heuristic approach to adapt the greediness of power mean.
Moreover, we are interested in analysing the bias and variance of the power mean estimator, or analyse the regret bound of \alg~ in MCTS.
Furthermore, we plan to test our methodology in more challenging Reinforcement Learning problems through the use of parametric function approximators, e.g. neural networks.

\section*{Acknowledgements}
This project has received funding from SKILLS4ROBOTS, project reference \#640554, and, by the German Research Foundation project PA 3179/1-1 (ROBOLEAP), and from the European Union\text{’}s Horizon 2020 research and innovation programme under grant agreement No. \#713010 (GOAL-Robots).

\clearpage
\bibliographystyle{named}
\bibliography{GeneralizedFMean}

\begin{thebibliography}{}

\bibitem[\protect\citeauthoryear{Auer \bgroup \em et al.\egroup
  }{2002}]{auer2002finite}
Peter Auer, Nicolo Cesa-Bianchi, and Paul Fischer.
\newblock Finite-time analysis of the multiarmed bandit problem.
\newblock {\em Machine learning}, 47(2-3):235--256, 2002.

\bibitem[\protect\citeauthoryear{Brockman \bgroup \em et al.\egroup
  }{2016}]{brockman2016openai}
Greg Brockman, Vicki Cheung, Ludwig Pettersson, Jonas Schneider, John Schulman,
  Jie Tang, and Wojciech Zaremba.
\newblock Openai gym.
\newblock {\em arXiv preprint arXiv:1606.01540}, 2016.

\bibitem[\protect\citeauthoryear{Bullen}{2013}]{bullen2013handbook}
Peter~S Bullen.
\newblock {\em Handbook of means and their inequalities}.
\newblock Springer Science \& Business Media, 2013.

\bibitem[\protect\citeauthoryear{Chaslot \bgroup \em et al.\egroup
  }{2008}]{chaslot2008progressive}
Guillaume Chaslot, Mark Winands, Jaap Van~Den Herik, Jos Uiterwijk, and Bruno
  Bouzy.
\newblock Progressive strategies for monte-carlo tree search.
\newblock {\em New Mathematics and Natural Computation}, 4(03):343--357, 2008.

\bibitem[\protect\citeauthoryear{Childs \bgroup \em et al.\egroup
  }{2008}]{childs2008transpositions}
Benjamin~E Childs, James~H Brodeur, and Levente Kocsis.
\newblock Transpositions and move groups in monte carlo tree search.
\newblock In {\em 2008 IEEE Symposium On Computational Intelligence and Games}.
  IEEE, 2008.

\bibitem[\protect\citeauthoryear{Coulom}{2006}]{coulom2006efficient}
R{\'e}mi Coulom.
\newblock Efficient selectivity and backup operators in monte-carlo tree
  search.
\newblock In {\em International conference on computers and games}. Springer,
  2006.

\bibitem[\protect\citeauthoryear{Dam \bgroup \em et al.\egroup
  }{2019}]{dam2019generalized}
Tuan Dam, Pascal Klink, Carlo D'Eramo, Jan Peters, and Joni Pajarinen.
\newblock Generalized mean estimation in monte-carlo tree search, 2019.

\bibitem[\protect\citeauthoryear{D'Eramo \bgroup \em et al.\egroup
  }{2016}]{d2016estimating}
Carlo D'Eramo, Marcello Restelli, and Alessandro Nuara.
\newblock Estimating maximum expected value through gaussian approximation.
\newblock In {\em International Conference on Machine Learning}, 2016.

\bibitem[\protect\citeauthoryear{Gelly and Silver}{2007}]{gelly2007combining}
Sylvain Gelly and David Silver.
\newblock Combining online and offline knowledge in uct.
\newblock In {\em Proceedings of the 24th international conference on Machine
  learning}, pages 273--280. ACM, 2007.

\bibitem[\protect\citeauthoryear{Gelly and Wang}{2006}]{gelly2006exploration}
Sylvain Gelly and Yizao Wang.
\newblock Exploration exploitation in go: Uct for monte-carlo go.
\newblock In {\em NIPS: Neural Information Processing Systems Conference
  On-line trading of Exploration and Exploitation Workshop}, 2006.

\bibitem[\protect\citeauthoryear{Hasselt}{2010}]{hasselt2010double}
Hado~V Hasselt.
\newblock Double q-learning.
\newblock In {\em Advances in Neural Information Processing Systems}, 2010.

\bibitem[\protect\citeauthoryear{Helmbold and
  Parker-Wood}{2009}]{helmbold2009all}
David~P Helmbold and Aleatha Parker-Wood.
\newblock All-moves-as-first heuristics in monte-carlo go.
\newblock In {\em IC-AI}, pages 605--610, 2009.

\bibitem[\protect\citeauthoryear{Hoock \bgroup \em et al.\egroup
  }{2010}]{hoock2010intelligent}
Jean-Baptiste Hoock, Chang-Shing Lee, Arpad Rimmel, Fabien Teytaud, Mei-Hui
  Wang, and Oliver Teytaud.
\newblock Intelligent agents for the game of go.
\newblock {\em IEEE Computational Intelligence Magazine}, 2010.

\bibitem[\protect\citeauthoryear{Khandelwal \bgroup \em et al.\egroup
  }{2016}]{khandelwal2016analysis}
Piyush Khandelwal, Elad Liebman, Scott Niekum, and Peter Stone.
\newblock On the analysis of complex backup strategies in monte carlo tree
  search.
\newblock In {\em International Conference on Machine Learning}, 2016.

\bibitem[\protect\citeauthoryear{Kocsis \bgroup \em et al.\egroup
  }{2006}]{kocsis2006improved}
Levente Kocsis, Csaba Szepesv{\'a}ri, and Jan Willemson.
\newblock Improved monte-carlo search.
\newblock 2006.

\bibitem[\protect\citeauthoryear{Kozelek}{2009}]{kozelek2009methods}
Tom{\'a}{\v{s}} Kozelek.
\newblock Methods of mcts and the game arimaa.
\newblock 2009.

\bibitem[\protect\citeauthoryear{Lorentz}{2010}]{lorentz2010improving}
Richard~J Lorentz.
\newblock Improving monte--carlo tree search in havannah.
\newblock In {\em International Conference on Computers and Games}, pages
  105--115. Springer, 2010.

\bibitem[\protect\citeauthoryear{Mansley \bgroup \em et al.\egroup
  }{2011}]{mansley2011sample}
Chris Mansley, Ari Weinstein, and Michael Littman.
\newblock Sample-based planning for continuous action markov decision
  processesgeneralizedfmean.
\newblock In {\em Twenty-First International Conference on Automated Planning
  and Scheduling}, 2011.

\bibitem[\protect\citeauthoryear{Mitrinovic and
  Vasic}{1970}]{mitrinovic1970analytic}
Dragoslav~S Mitrinovic and Petar~M Vasic.
\newblock {\em Analytic inequalities}.
\newblock Springer, 1970.

\bibitem[\protect\citeauthoryear{Rummery}{1995}]{rummery1995problem}
Gavin~Adrian Rummery.
\newblock {\em Problem solving with reinforcement learning}.
\newblock PhD thesis, University of Cambridge Ph. D. dissertation, 1995.

\bibitem[\protect\citeauthoryear{Silver and Veness}{2010}]{silver2010monte}
David Silver and Joel Veness.
\newblock Monte-carlo planning in large pomdps.
\newblock In {\em Advances in neural information processing systems}, 2010.

\bibitem[\protect\citeauthoryear{Silver \bgroup \em et al.\egroup
  }{2016}]{silver2016mastering}
David Silver, Aja Huang, Chris~J Maddison, Arthur Guez, Laurent Sifre, George
  Van Den~Driessche, Julian Schrittwieser, Ioannis Antonoglou, Veda
  Panneershelvam, Marc Lanctot, et~al.
\newblock Mastering the game of go with deep neural networks and tree search.
\newblock {\em nature}, 529(7587):484, 2016.

\bibitem[\protect\citeauthoryear{Smith and Simmons}{2004}]{smith2004heuristic}
Trey Smith and Reid Simmons.
\newblock Heuristic search value iteration for pomdps.
\newblock In {\em Proceedings of the 20th conference on Uncertainty in
  artificial intelligence}, pages 520--527. AUAI Press, 2004.

\bibitem[\protect\citeauthoryear{Smith and Winkler}{2006}]{smith2006optimizer}
James~E Smith and Robert~L Winkler.
\newblock The optimizer’s curse: Skepticism and postdecision surprise in
  decision analysis.
\newblock {\em Management Science}, 52(3), 2006.

\bibitem[\protect\citeauthoryear{Tesauro \bgroup \em et al.\egroup
  }{2012}]{tesauro2012bayesian}
Gerald Tesauro, VT~Rajan, and Richard Segal.
\newblock Bayesian inference in monte-carlo tree search.
\newblock {\em arXiv preprint arXiv:1203.3519}, 2012.

\bibitem[\protect\citeauthoryear{Teytaud and Teytaud}{2010}]{teytaud2010huge}
Fabien Teytaud and Olivier Teytaud.
\newblock On the huge benefit of decisive moves in monte-carlo tree search
  algorithms.
\newblock In {\em Proceedings of the 2010 IEEE Conference on Computational
  Intelligence and Games}. IEEE, 2010.

\bibitem[\protect\citeauthoryear{Tom}{2010}]{tom2010investigating}
David Tom.
\newblock Investigating uct and rave: Steps towards a more robust method.
\newblock 2010.

\bibitem[\protect\citeauthoryear{Vodopivec \bgroup \em et al.\egroup
  }{2017}]{vodopivec2017monte}
Tom Vodopivec, Spyridon Samothrakis, and Branko Ster.
\newblock On monte carlo tree search and reinforcement learning.
\newblock {\em Journal of Artificial Intelligence Research}, 60:881--936, 2017.

\bibitem[\protect\citeauthoryear{Wasserman}{2004}]{wasserman1all}
Larry Wasserman.
\newblock All of statistics: a concise course in statistical inference. 2004,
  2004.

\bibitem[\protect\citeauthoryear{Watkins}{1989}]{watkins:phd}
C.~Watkins.
\newblock {\em Learning from Delayed Rewards}.
\newblock PhD thesis, King's College, Cambridge, England, 1989.

\bibitem[\protect\citeauthoryear{Xiao \bgroup \em et al.\egroup
  }{2019}]{xiao2019maximum}
Chenjun Xiao, Ruitong Huang, Jincheng Mei, Dale Schuurmans, and Martin
  M{\"u}ller.
\newblock Maximum entropy monte-carlo planning.
\newblock In {\em Advances in Neural Information Processing Systems}, pages
  9516--9524, 2019.

\end{thebibliography}

\onecolumn
\appendix
\section{Appendix}
We derive here the proof of convergence for \alg. The proof is based on the proof of the UCT~\cite{kocsis2006improved} method but differs in several key places.
In this section, we show that \alg~ can smoothly adapt to all theorems of UCT~\cite{kocsis2006improved}. The following results can be seen as a generalization of the results for UCT, as we consider a generalized mean instead of a standard mean as the backup operator.
Our main results are Theorem 6 and Theorem 7, which respectively prove the convergence of failure probability at the root node, and derive the bias of power mean estimated payoff. 
In order to prove them, we start with Theorem \ref{theorem1} to show the concentration of power mean with respect to i.i.d random variable $X$. Subsequently, Theorem 2 shows the upper bound of the expected number of times when a suboptimal arm is played.
Theorem 3 bounds the expected error of the power mean estimation.
Theorem 4 shows the lower bound of the number of times any arm is played.
Theorem 5 shows the concentration of power mean backup around its mean value at each node in the tree. 

\noindent We start with well-known lemmas and respective proofs:
The following lemma shows that Power Mean can be upper bound by Average Mean plus with a constant
\begin{manuallemma}{2}\label{lb:lemma2}
Let $0 < l \leq X_i \leq U, C = \frac{U}{l}, \forall i \in (1, ..., n) $ and $p > q$. We define:
\begin{flalign*}
\text{Q}(X, w, p, q) &= \frac{\text{M}^{[p]}_n(X,w)}{\text{M}^{[q]}_n(X,w)}\\
\text{D}(X, w, p, q) &= \text{M}^{[p]}_n(X,w) - \text{M}^{[q]}_n(X,w).\\
\end{flalign*}

Then we have:
\begin{flalign}
&\text{Q}(X, w, p, q) \leq \text{L}_{p,q} \nonumber \\
&\text{D}(X, w, p, q) \leq \text{H}_{p,q} \nonumber \\
&\text{L}_{p,q} = \Bigg( \frac{q(C^p - C^q)}{(p-q)(C^q - 1)}\bigg)^{\frac{1}{p}} 
\Bigg( \frac{p(C^q - C^p)}{(q-p)(C^p - 1)} \bigg)^{-\frac{1}{q}} \nonumber \\
&\text{H}_{p,q} = (\theta U^p + (1 - \theta) l^p)^{\frac{1}{p}} - (\theta U^q + (1 - \theta) l^q)^{1/q}, \nonumber
\end{flalign}

\noindent where $\theta$ is defined in the following way. Let\\
\begin{flalign}
h(x) &= x^{\frac{1}{p}} - (ax + b)^{1/q} \nonumber
\end{flalign}
where:
\begin{flalign}
a &= \frac{U^q - l^q}{U^p - l^p} \nonumber\\
b &= \frac{U^p l^q - U^q l^p}{U^p - l^p} \nonumber\\
x^{'} &= \argmax \{h(x), x \in (l^p, U^p)\}\nonumber
\end{flalign}
then:
\begin{flalign}
\theta = \frac{x' - l^p}{U^p - l^p}.\nonumber
\end{flalign}
\end{manuallemma}
\begin{proof}
Refer to \citet{mitrinovic1970analytic}.
\end{proof}{}

\begin{manuallemma}{3}
Let $X$ be an independent random variable with common mean $\mu$ and $a \leq X \leq b$. Then for any t\\
\begin{flalign}
\E[\exp(tX)] \leq \exp\left(t\mu + t^2 \frac{(b-a)^2}{8}\right)
\end{flalign}
\end{manuallemma}

\begin{proof}
Refer to \citet{wasserman1all} page 67.
\end{proof}
\begin{manuallemma}{4}
\text{Chernoff’s inequality} $t > 0$,
\begin{flalign}
\Pr (X > \epsilon) \leq \exp(-t\epsilon) \E [\exp(tX)]
\end{flalign}
\end{manuallemma}

\begin{proof}
This is a well-known result.
\end{proof}

\noindent The next result show the generalization of Hoeffding Inequality for Power Mean estimation

\begin{manualtheorem}{1}\label{theorem1}
If $X_1, X_2, ..., X_n$ are independent with $\Pr(a < X_i \leq b) = 1$
and common mean $\mu$, $w_1, w_2, ..., w_n$ are positive and $W = \sum^n_{i=1} w_i$ then for any $\epsilon > 0$, $p \geq 1$
\begin{flalign}
\Pr \left( \left| \left( \frac{\sum^n_{i=1} w_i X_i^p}{\sum^n_{i=1} w_i} \right)^{\frac{1}{p}} - \mu \right| > \epsilon \right) \nonumber \\
\leq 2 \exp\left(\text{H}_{p, 1}\right)\exp\left(-\frac{2 \epsilon^2 W^2}{(b-a)^2 \sum^n_{i=1} w_i^2}\right) \nonumber
\end{flalign}
\end{manualtheorem}

\begin{proof}
We have
\begin{flalign}
&\Pr \left ( \left( \frac{\sum^n_{i=1} w_i X_i^p}{W} \right)^{\frac{1}{p}} - \mu > \epsilon \right)\nonumber \\
&\leq \exp(-t\epsilon) \E\left[ \exp\left(\left(t\frac{\sum^n_{i=1} w_i X_i^p}{W}\right)^{\frac{1}{p}} - t\mu\right)\right] \text {( see Lemma 4)} \nonumber \\
&\leq \exp(-t\epsilon) \E\left[ \exp\left(\left(t\frac{\sum^n_{i=1} w_i X_i}{W}\right) - t\mu + \text{H}_{p,1}\right)\right] \text {( see Lemma 2)} \nonumber \\
&=\exp(\text{H}_{p,1})\exp(-t\epsilon) \E\left[ \exp\left(t\frac{\sum^n_{i=1} w_i (X_i - \mu)}{W}\right)\right]\nonumber \\
&=\exp(\text{H}_{p,1})\exp(-t\epsilon) \prod_{i=1}^n \E\left[ \exp\left(t \frac{w_i }{W} (X_i - \mu)\right)\right] \nonumber \\
&\leq\exp(\text{H}_{p,1})\exp(-t\epsilon) \prod_{i=1}^n \exp\left(\left(t\frac{w_i}{W}\right)^2 \frac{(b - a)^2}{8}\right) \text {( see Lemma 3)} \nonumber \\
&= \exp(\text{H}_{p,1})\exp\left(t^2 \frac{(b - a)^2 \sum^n_{i=1} w_i^2}{W^2 8} -t\epsilon\right) \nonumber
\end{flalign}
and setting $t = \frac{4\epsilon W^2}{(b-a)^2 \sum^n_{i=1} w_i^2}$ yields
\begin{flalign}
&\Pr \left( \left( \frac{\sum^n_{i=1} w_i X_i^p}{\sum^n_{i=1} w_i} \right)^{\frac{1}{p}} - \mu > \epsilon \right) \leq \exp(\text{H}_{p,1})\exp\left(- \frac{2 \epsilon^2 W^2}{(b-a)^2 \sum^n_{i=1} w_i^2}\right). \nonumber
\end{flalign}


\noindent Similarly, for $p \geq 1$, Power Mean is always greater than Mean. Hence, the inequality holds for $\exp(H_{1,1})$ which is 1
\begin{flalign}
&\Pr \left ( \left( \frac{\sum^n_{i=1} w_i X_i^p}{\sum^n_{i=1} w_i} \right)^{\frac{1}{p}} - \mu < -\epsilon \right) \leq \exp\left(- \frac{2 \epsilon^2 W^2}{(b-a)^2 \sum^n_{i=1} w_i^2}\right) \nonumber
\end{flalign}

\noindent which completes the proof.
\end{proof}

For the following proofs, we will define a special case of this inequality. Setting a = 0, b = 1, $w_i$ = 1 $(\forall i = 1, 2...n)$ yields
\begin{flalign}
&\Pr \left ( \left( \frac{\sum^n_{i=1} X_i^p}{n} \right)^{\frac{1}{p}} - \mu > \epsilon \right) \leq \exp(H_{p,1}) \exp\left(-2n \epsilon^2\right) \nonumber\\
&\Pr \left ( \left( \frac{\sum^n_{i=1} X_i^p}{n} \right)^{\frac{1}{p}} - \mu < -\epsilon \right) \leq \exp\left(-2n \epsilon^2\right) \nonumber
\end{flalign}
With $\triangle_n = 9\sqrt{2n\log(\frac{2M}{\delta})}$, ($M > 0$ and $\delta > 0$ are constant) we get
\begin{flalign}
&\Pr \left ( \left( \frac{\sum^n_{i=1} X_i^p}{n} \right)^{\frac{1}{p}} - \mu > \frac{\triangle_n}{9n} \right) \leq \exp(H_{p,1}) \exp\left(-2n \left(\frac{\triangle_n}{9n}\right)^2\right). \nonumber
\end{flalign}
\noindent Therefore,\\
\begin{flalign}
&\Pr \left ( \left( \frac{\sum^n_{i=1} X_i^p}{n} \right)^{\frac{1}{p}} - \mu > \frac{\triangle_n}{9n} \right) \leq \exp(H_{p,1}) \exp\left( -4 \log\left( \frac{2M}{\delta} \right) \right) = \exp(H_{p,1}) \left( \frac{\delta}{2M} \right)^4.
\end{flalign}
Due to the definition of $\triangle_n$, we only need to consider the case $\frac{\delta}{2M} <= 1$,  since for the case $\frac{\delta}{2M} > 1$ it follows that $\frac{2M}{\delta} < 1$ and hence the $log$-term in $\triangle_n$ would become negative. With this additional information, we can further bound above probability
\begin{flalign}
&\Pr \left ( \left( \frac{\sum^n_{i=1} X_i^p}{n} \right)^{\frac{1}{p}} - \mu > \frac{\triangle_n}{9n} \right) \leq \exp(H_{p,1}) \left( \frac{\delta}{2M} \right)^4 \leq \exp(H_{p,1}) \frac{\delta}{2M}.
\end{flalign}
If $M=\exp(H_{p,1})$ where $H_{p,1}$ is defined in Lemma \ref{lb:lemma2}, we have
\begin{flalign}
&\Pr \left ( \left( \frac{\sum^n_{i=1} X_i^p}{n} \right)^{\frac{1}{p}} - \mu > \frac{\triangle_n}{9n} \right) \leq \frac{\delta}{2} \label{19}.
\end{flalign}


\noindent Let's start with an assumption:
\begin{manualassumption}{1}\label{asumpt}
Fix $1 \leq i \leq K$. Let $\{F_{it}\}_t$ be a filtration such that$ \{X_{it}\}_t$ is $\{F_{it}\}$-adapted
and $X_{i,t}$ is conditionally independent of $F_{i,t+1}, F_{i,t+2},...$ given $F_{i,t-1}$. Then $0 \leq X_{it} \leq 1$ and the limit of $\mu_{in} = \E[\overline{X_{in}}(p)]$ exists, Further, we assume that there exists a constant $C > 0$ and an integer $N_c$ such that for $n>N_c$, for any $\delta > 0$, $ \triangle_n(\delta) = C\sqrt{n\log(1/\delta)}$, the following bounds hold:
\begin{flalign}
\Pr(\overline{X}_{in}(p) \geq \E[ \overline{X}_{in}(p)] + \triangle_n(\delta)/n) \leq \delta \label{eq:3}, \\
\Pr(\overline{X}_{in}(p) \leq \E[ \overline{X}_{in}(p)] - \triangle_n(\delta)/n) \leq \delta \label{eq:4}.
\end{flalign}
\end{manualassumption}

\noindent Under Assumption 1, For any internal node arm $k$, at time step $t$, let define $\mu_{kt}=\E[ \overline{X}_{kt}(p)]$, a suitable choice for bias sequence is that $c_{t,s} = 2C\sqrt{\frac{\log t}{s}}$ (C is an exploration constant) used in UCB1 (using power mean estimator), we get
\begin{flalign}
&\Pr \Bigg ( \Big( \frac{\sum^{s}_{i=1} X_{ki}^p}{s} \Big)^{\frac{1}{p}} - \mu_{kt} > 2C\sqrt{\frac{\log t}{s}} \Bigg) \leq t^{-4} \label{21} \\
&\Pr \Bigg ( \Big( \frac{\sum^{s}_{i=1} X_{ki}^p}{s} \Big)^{\frac{1}{p}} - \mu_{kt} < -2C\sqrt{\frac{\log t}{s}} \Bigg) \leq t^{-4}. \label{22} 
\end{flalign}

\noindent From Assumption \ref{asumpt}, we derive the upper bound for the expectation of the number of plays a sub-optimal arm:
\begin{manualtheorem} {2}\label{T:th_2}
Consider UCB1 (using power mean estimator) applied to a non-stationary problem where the pay-off sequence satisfies Assumption 1 and 
where the bias sequence, $c_{t,s} = 2C\sqrt{\log t/s}$ (C is an exploration constant). Fix $\epsilon \geq 0$. Let $T_k(n)$ denote the number of plays of arm $k$. Then if $k$ is the index of a suboptimal arm then
Each sub-optimal arm $k$ is played in expectation at most
\begin{flalign}
\E[T_k(n)] \leq \frac{16C^2\ln n}{(1-\epsilon)^2 \triangle_k^2} + A(\epsilon) + N_c + \frac{\pi^2}{3} + 1.
\end{flalign}
\end{manualtheorem}
\begin{proof}
When a sub-optimal arm $k$ is pulled at time $t$ we get
\begin{flalign}
\Bigg( \frac{\sum^{T_k(t-1)}_{i=1} X_{k,i}^p}{T_k(t-1)} \Bigg)^{\frac{1}{p}} + 2C\sqrt{\frac{\ln t}{T_k(t-1)}} \geq \Bigg( \frac{\sum^{T_{k^*}(t-1)}_{i=1} X_{k^*,i}^p}{T_{k^*}(t-1)} \Bigg)^{\frac{1}{p}} + 2C\sqrt{\frac{\ln t}{T_{k^*}(t-1)}} \label{lb_pull_arm_k}
\end{flalign}
Now, consider the following two inequalities:
\begin{itemize}
    \item The empirical mean of the optimal arm is not within its confidence interval:
    \begin{flalign}
    \Bigg( \frac{\sum^{T_{k^*}(t-1)}_{i=1} X_{k^*,i}^p}{T_{k^*}(t-1)} \Bigg)^{\frac{1}{p}} + 2C\sqrt{\frac{\ln t}{T_{k^*}(t-1)}} \leq \mu_t^* \label{lb_optimal}
    \end{flalign}
    \item The empirical mean of the arm k is not within its confidence interval:
    \begin{flalign}
    \Bigg( \frac{\sum^{T_k(t-1)}_{i=1} X_{k,i}^p}{T_k(t-1)} \Bigg)^{\frac{1}{p}} \geq \mu_{kt}+  2C\sqrt{\frac{\ln t}{T_k(t-1)}} \label{lb_arm_k}
    \end{flalign}
\end{itemize}
If both previous inequalities (\ref{lb_optimal}), (\ref{lb_arm_k}) do not hold, and if a sub-optimal arm k is pulled, then we deduce that
\begin{flalign}
    \mu_{kt}+  2C\sqrt{\frac{\ln t}{T_k(t-1)}} \geq \Bigg( \frac{\sum^{T_k(t-1)}_{i=1} X_{k,i}^p}{T_k(t-1)} \Bigg)^{\frac{1}{p}} \text{ see } (\ref{lb_arm_k})
\end{flalign}
and
\begin{flalign}
\Bigg( \frac{\sum^{T_k(t-1)}_{i=1} X_{k,i}^p}{T_k(t-1)} \Bigg)^{\frac{1}{p}} \geq \Bigg( \frac{\sum^{T_{k^*}(t-1)}_{i=1} X_{k^*,i}^p}{T_{k^*}(t-1)} \Bigg)^{\frac{1}{p}} + 2C\sqrt{\frac{\ln t}{T_{k^*}(t-1)}} - 2C\sqrt{\frac{\ln t}{T_k(t-1)}} \text{ see } (\ref{lb_pull_arm_k})
\end{flalign}
and
\begin{flalign}
    \Bigg( \frac{\sum^{T_{k^*}(t-1)}_{i=1} X_{k^*,i}^p}{T_{k^*}(t-1)} \Bigg)^{\frac{1}{p}} + 2C\sqrt{\frac{\ln t}{T_{k^*}(t-1)}} \geq \mu_t^* \text{ see } (\ref{lb_optimal}).
\end{flalign}
\noindent So that
\begin{flalign}
\mu_{kt} + 4C\sqrt{\frac{\ln t}{T_k(t-1)}} \geq \mu_t^*.
\end{flalign}
$\mu_{kt} = \mu_k + \delta_{kt}$, $\mu_t^{*} = \mu^{*} + \delta_{t}^{*}$ and we have an assumption that $\lim_{t\rightarrow \infty}\mu_{kt} = \mu_k$ for any $k \in [1,2,...K]$ yields $\lim_{t\rightarrow \infty}\delta_{kt} = 0$
Therefore, for any $\epsilon > 0$, we can find an index $A(\epsilon)$ such that for any $t > A(\epsilon)$:
$\delta_{kt} \leq \epsilon \triangle_k$ with $\triangle_k = \mu^* - \mu_k$. Which means that
\begin{flalign}
4C\sqrt{\frac{\ln t}{T_k(t-1)}} \geq \triangle_k - \delta_{kt} + \delta_{t}^{*} \geq (1-\epsilon)\triangle_k
\end{flalign}
\noindent which implies $T_k(t-1) \leq \frac{16C^2\ln t}{(1-\epsilon)^2 \triangle_k^2}$.\\
This says that whenever $T_k(t-1) \geq \frac{16C^2\ln t}{(1-\epsilon)^2 \triangle_k^2} + A(\epsilon) + N_c$, either arm $k$ is not pulled at time t or one of the two following events (\ref{lb_optimal}), (\ref{lb_arm_k}) holds. Thus if we define $u = \frac{16C^2\ln t}{(1-\epsilon)^2 \triangle_k^2} + A(\epsilon) + N_c$, we have:
\begin{flalign}
T_k(n) &\leq u + \sum_{t=u+1}^{n} \textbf{1}\{I_t=k; T_k(t) \geq u\} \nonumber  \\
      &\leq u + \sum_{t=u+1}^{n} \textbf{1}\{(\ref{lb_optimal}), \text{ or } (\ref{lb_arm_k}) \text{ holds }\} \label{lb_60_61} \nonumber 
\end{flalign}
\noindent We have from (\ref{21}),(\ref{22}):
\begin{flalign}
\Pr\Bigg( \Big( \frac{\sum^{T_{k^*}(t-1)}_{i=1} X_{k^*,i}^p}{T_{k^*}(t-1)} \Big)^{\frac{1}{p}} + 2C\sqrt{\frac{\ln t}{T_{k^*}(t-1)}} \leq \mu_t^* \Bigg) \leq \sum_{s=1}^{t} \frac{1}{t^{4}} = \frac{1}{t^{3}}
\end{flalign}
\noindent and:
\begin{flalign}
\Pr\Bigg( \Big( \frac{\sum^{T_k(t-1)}_{i=1} X_{k,i}^p}{T_k(t-1)} \Big)^{\frac{1}{p}} \geq \mu_{kt}+  2C\sqrt{\frac{\ln t}{T_k(t-1)}} \Bigg) \leq \sum_{s=1}^{t} \frac{1}{t^{4}} = \frac{1}{t^{3}}
\end{flalign}
\noindent so that from (\ref{lb_60_61}), we have
\begin{flalign}
\E[T_k(n)] &\leq \frac{16C^2\ln t}{(1-\epsilon)^2 \triangle_k^2} + A(\epsilon) + N_c + \sum_{t=u+1}^{n}\frac{2}{t^{8C^2-1}} = \frac{16C^2\ln t}{(1-\epsilon)^2 \triangle_k^2} + A(\epsilon) + N_c + + \sum_{t=u+1}^{n}\frac{2}{t^3} \nonumber \\
&\leq \frac{16C^2\ln t}{(1-\epsilon)^2 \triangle_k^2} + A(\epsilon) + N_c + \frac{\pi^2}{3} \nonumber 
\end{flalign}
\end{proof}
\noindent Based on this result we derive an upper bound for the expectation of power mean in the next theorem as follows.

\begin{manualtheorem} {3}
Under the assumptions of Theorem~\ref{T:th_2},
\begin{flalign}
\big| \E\big[ \overline{X}_n(p) \big]  - \mu^{*} \big| &\leq |\delta^*_n| + \mathcal{O} \Bigg( \frac{K(C^2 \log n + N_0)}{n} \Bigg)^{\frac{1}{p}}. \nonumber
\end{flalign}
\end{manualtheorem}
\begin{proof}
In UCT, the value of each node is used for backup as $\overline{X}_n = \sum^{K}_{i=1} \left(\frac{T_i(n)}{n}\right) \overline{X}_{i, T_i(n)}$, and the authors show that 
\begin{flalign}
&\big| \E\big[ \overline{X}_n \big]  - \mu^{*} \big| \leq \big| \E\big[ \overline{X}_n \big]  - \mu^{*}_{n} \big| + \big| \mu^{*}_{n}  - \mu^{*} \big|\nonumber \\
&= \big| \delta^{*}_{n} \big| + \big| \E\big[ \overline{X}_n \big]  - \mu^{*}_{n} \big| \nonumber \\
\label{11} &\leq \big| \delta^{*}_{n} \big| + \mathcal{O} \Bigg( \frac{K(C^2 \log n + N_0)}{n} \Bigg)
\end{flalign}
We derive the same results replacing the average with the power mean. First, we have
\begin{flalign}
&\E\left[ \overline{X}_n(p) \right] - \mu_{n}^{*} = \E\left[ \left(\sum^{K}_{i=1} \frac{T_i(n)}{n} \overline{X}_{i, T_i(n)}^p\right)^{\frac{1}{p}} \right] - \mu_{n}^{*}.
\end{flalign}
In the proof, we will make use of the following inequalities:
\begin{flalign}
\label{reward_01} &0 \leq X_i \leq 1, \\
\label{13} &x^{\frac{1}{p}} \leq y^{\frac{1}{p}} \space \text{ when 0 } \leq {x} \leq { y }, \\
\label{14} &(x + y)^m \leq x^m + y^m \space (0 \leq m \leq 1), \\
\label{15} &\E[f(X)] \leq f(\E[X]) \textit{ (f(X)} \text{ is concave)}.
\end{flalign}
With $i^*$ being the index of the optimal arm, we can derive an upper bound on the difference between the value backup and the true average reward:
\begin{flalign}
&\E\left[ \left(\sum^{K}_{i=1} \frac{T_i(n)}{n} \overline{X}_{i, T_i(n)}^p\right)^{\frac{1}{p}} \right] - \mu_n^{*} \nonumber\\
&\leq \E\left[ \left( \left(\sum^{K}_{i=1;i\neq i^{*}} \frac{T_i(n)}{n} \right) + \overline{X}_{i*, T_i*(n)}^p\right)^{\frac{1}{p}} \right] - \mu_n^{*} (\text{see } (\ref{reward_01})) \nonumber\\
&\leq \E\left[ \left(\sum^{K}_{i=1;i\neq i^{*}} \frac{T_i(n)}{n} \right)^{\frac{1}{p}} + \overline{X}_{i*, T_i*(n)} \right] - \mu_n^{*} (\text{see } (\ref{14}))\nonumber\\
&= \E\left[ \left(\sum^{K}_{i=1;i\neq i^{*}} \frac{T_i(n)}{n} \right)^{\frac{1}{p}} \right] + \E \left[\overline{X}_{i*, T_i*(n)}\right]  - \mu_n^{*}\nonumber\\
&= \E\left[ \left(\sum^{K}_{i=1;i\neq i^{*}} \frac{T_i(n)}{n} \right)^{\frac{1}{p}} \right] \nonumber\\
&\leq \left(\sum^{K}_{i=1;i\neq i^{*}} \E \left[\frac{T_i(n)}{n}\right] \right)^{\frac{1}{p}} (\text{see } (\ref{15})) \nonumber\\
\label{16} &\leq ( ( K-1 )\mathcal{O} \Bigg( \frac{K(C^2 \log{n} + N_0)}{n} \Bigg) )^{\frac{1}{p}} (\text{Theorem~\ref{T:th_2} \& } (\ref{13}))
\end{flalign}
According to Lemma 1, it holds that
$$
E\left[\overline{X}_n(p)\right] \geq E\left[\overline{X}_n\right]
$$
for $p \geq 1$. Because of this, we can reuse the lower bound given by (\ref{11}):
\begin{flalign}
& -\mathcal{O} \Bigg( \frac{K(C^2 \log n + N_0)}{n} \Bigg) \leq E\big[ \overline{X}_n \big] - \mu_{n}^{*},\nonumber
\end{flalign}
so that: 
\begin{flalign}
& -\mathcal{O} \Bigg( \frac{K(C^2 \log n + N_0)}{n} \Bigg) \leq \E\big[ \overline{X}_n \big] - \mu_{n}^{*}\nonumber\\
\label{17} &\leq \E\big[\overline{X}_n(p) \big] - \mu_{n}^{*}.
\end{flalign}
Combining (\ref{16}) and (\ref{17}) concludes our prove:
\begin{flalign}
\big| \E\big[ \overline{X}_n(p) \big]  - \mu^{*} \big| &\leq |\delta^*_n| + O \Bigg( \frac{K(C^2 \log n + N_0)}{n} \Bigg)^{\frac{1}{p}}. \nonumber
\end{flalign}
\end{proof}

\noindent The following theorem shows lower bound of choosing all the arms:
\begin{manualtheorem} {4} (\textbf{Lower Bound})
Under the assumptions of Theorem 2, there exists some positive constant $\rho$ such that for all arms k and n,
$T_k(n) \geq \lceil \rho \log (n)\rceil$
\end{manualtheorem}
\begin{proof}
There should exist a constant $S$ that
\begin{flalign}
\Bigg( \frac{\sum^{T_k(t-1)}_{i=1} X_{k,i}^p}{T_k(t-1)} \Bigg)^{\frac{1}{p}} + 2C\sqrt{\frac{\ln t}{T_k(t-1)}} \leq S \nonumber 
\end{flalign}
for all arm k so
\begin{flalign}
\mu_{k} + \delta_{kt} + 2C\sqrt{\frac{\log t}{T_k(t-1)}} \leq S \nonumber 
\end{flalign}
because
\begin{flalign}
\lim_{t\rightarrow \infty} \delta_{kt} = 0\nonumber
\end{flalign}
so there exists a positive constant $\rho$ that $T_k(t-1) \geq \lceil \rho \log (t)\rceil$
\end{proof}

\noindent The next result shows the estimated optimal payoff concentration around its mean (Theorem \ref{theorem5}).
In order to prove that, \noindent we now reproduce here Lemma \ref{lb:lemma5}, \ref{lb:lemma6} ~\cite{kocsis2006improved} that we use for our proof:



\begin{manuallemma}{5} \textbf{Hoeffding-Azuma inequality for Stopped Martingales} (Lemma 10 in~\cite{kocsis2006improved}). \label{lb:lemma5}
Assume that $S_t$ is a centered martingale such that the corresponding martingale difference
process is uniformly bounded by C. Then, for any fixed $\epsilon \geq 0$, integers $0 \leq a \leq b$, the following inequalities hold:
\begin{flalign}
\Pr(S_N \geq \epsilon N) \leq (b-a+1) \exp\Big(\frac{-2a^2 \epsilon^2}{C^2}\Big) + \Pr(N \notin [a,b]),\\
\Pr(S_N \leq \epsilon N) \leq (b-a+1) \exp\Big(\frac{-2a^2 \epsilon^2}{C^2}\Big) + \Pr(N \notin [a,b]),
\end{flalign}
\end{manuallemma}




\begin{manuallemma}{6} \label{lb:lemma6} (Lemma 13 in~\cite{kocsis2006improved})
Let ($Z_i$), i=1,...,n be a sequence of random variables such that $Z_i$ is conditionally independent of $Z_{i+1}, ...,Z_n$ given $Z_1, ...,Z_{i-1}$. Let define $N_n=\sum_{i=1}^n Z_i$, and let $a_n$ is an upper bound on $\E [N_n]$. Then for all $\triangle \geq 0$, if n is such that $a_n \leq \triangle/2$ then    
\begin{flalign}
\Pr(N_n \geq \triangle) \leq \exp(-\triangle^2/(8n)).
\end{flalign}
\end{manuallemma}

\noindent The next lemma is our core result for propagating confidence bounds upward in the tree, and it is used for the prove of Theorem 5 about the concentration of power mean estimator.

\begin{manuallemma}{7} \label{lb:lemma7}
let $Z_i$, $a_i$ be as in Lemma \ref{lb:lemma6}.
Let $F_i$ denotes a filtration over some probability space. $Y_i$ be an $F_i$-adapted real valued martingale-difference sequence. Let {$X_i$} be an i.i.d.\ sequence with mean 
$\mu$. We assume that both $X_i$ and $Y_i$ lie
in the [0,1] interval. Consider the partial sums\\
\begin{flalign}
S_n = \Bigg(\frac{\sum_{i=1}^n (1-Z_i) X_i^p + Z_i Y_i^p}{n}\Bigg)^{\frac{1}{p}}.
\end{flalign}
Fix an arbitrary $\delta > 0$, and fix $p \geq 1$, and $M = \exp(H_{p,1})$ where $H_{p,1}$ is defined as in Lemma \ref{lb:lemma2}. Let $\triangle_n = 9\sqrt{2n \log(2M/\delta)}$, and $\triangle = (9/4)^{p-1}\triangle_{n}$ let 
\begin{flalign}
R_n = \E\Bigg[\Bigg(\frac{\sum_{i = 1}^n X_i^p}{n}\Bigg)^{\frac{1}{p}}\Bigg] - \E[S_n] \label{23}.
\end{flalign}
Then for n such that $a_n \leq (1/9)\triangle_n$ and $|R_n| \leq (4/9) (\triangle/n)^{\frac{1}{p}}$
\begin{flalign}
\Pr(S_n \geq \E[S_n] + (\triangle/n)^{\frac{1}{p}}) \leq \delta \label{lb_lower}\\
\Pr(S_n \leq \E[S_n] - (\triangle/n)^{\frac{1}{p}}) \leq \delta \label{lb_upper}
\end{flalign}
\end{manuallemma}

\begin{proof}
We have a very fundamental probability inequality:\\
\noindent Consider two events: $A, B$. If $A \in B$, then $\Pr(A) \leq \Pr(B)$.\\
Therefore, if we have three random variables $X, Y, Z$ and if we are sure that 
\begin{flalign}
Y\geq Z, \text{ then } \Pr(X \geq Y) \leq \Pr(X \geq Z) \label{basic_pr} 
\end{flalign}
We have
\begin{flalign}
&\Bigg(\frac{\sum_{i=1}^n (1-Z_i) X_i^p + Z_i Y_i^p}{n}\Bigg)^{\frac{1}{p}} \nonumber \\ 
&= \Bigg(\frac{\sum_{i=1}^n X_i^p}{n} + \frac{Z_i(Y_i^p - X_i^p)}{n}\Bigg)^{\frac{1}{p}} \leq \Bigg(\frac{\sum_{i=1}^n X_i^p}{n} + \frac{2\sum_{i=1}^n Z_i}{n}\Bigg)^{\frac{1}{p}} (X_i, Y_i \in [0,1]) \nonumber \\
&\leq \Bigg(\frac{\sum_{i=1}^n X_i^p}{n}\Bigg)^{\frac{1}{p}} + \Bigg(\frac{2\sum_{i=1}^n Z_i}{n}\Bigg)^{\frac{1}{p}} (\text{see } (\ref{13})) \label{27}
\end{flalign}
Therefore,
\begin{flalign}
T &= \Pr\Bigg(S_n \geq \E[S_n] + (\triangle/n)^{\frac{1}{p}}\Bigg)\\ 
&= \Pr\Bigg(\Big(\frac{\sum_{i=1}^n (1-Z_i) X_i^p + Z_i Y_i^p}{n}\Big)^{\frac{1}{p}} \geq \E[\frac{\sum_{i = 1}^n X_i^p}{n}\Big)^{\frac{1}{p}}] - R_n + (\triangle/n)^{\frac{1}{p}}\Bigg) (\text{see } (\ref{23}))\\
&\leq \Pr\Bigg(\Big(\frac{\sum_{i=1}^n X_i^p}{n}\Big)^{\frac{1}{p}} + \Big(\frac{2\sum_{i=1}^n Z_i}{n}\Big)^{\frac{1}{p}} \geq \E\Bigg[\Big(\frac{\sum_{i = 1}^n X_i^p}{n}\Big)^{\frac{1}{p}}\Bigg] - R_n + (\triangle/n)^{\frac{1}{p}}\Bigg) (\text{see } (\ref{basic_pr}), (\ref{27}))) \nonumber
\end{flalign}
Using the elementary inequality $I(A+B\geq \triangle/n) \leq I(A\geq \alpha \triangle/n) + I(B\geq (1-\alpha)\triangle/n)$ that holds for any $A,B \geq 0; 0\leq \alpha \leq 1$, we get
\begin{flalign}
T &\leq \Pr\bigg(\Big(\frac{\sum_{i=1}^n X_i^p}{n}\Big)^{\frac{1}{p}} \geq E\Bigg[\Big(\frac{\sum_{i=1}^n X_i^p}{n}\Big)^{\frac{1}{p}}\Bigg] + 1/9(\triangle/n)^{\frac{1}{p}}\bigg) + \Pr\bigg(\Big(\frac{2\sum_{i=1}^n Z_i}{n}\Big)^{\frac{1}{p}} \geq 8/9(\triangle/n)^{\frac{1}{p}} - R_n\bigg) \nonumber\\
&\leq \Pr\bigg(\Big(\frac{\sum_{i=1}^n X_i^p}{n}\Big)^{\frac{1}{p}} \geq E\Bigg[\Big(\frac{\sum_{i=1}^n X_i}{n}\Big)\Bigg] + 1/9(\triangle/n)^{\frac{1}{p}}\bigg) + \Pr\bigg(\Big(\frac{2\sum_{i=1}^n Z_i}{n}\Big)^{\frac{1}{p}} \geq 8/9(\triangle/n)^{\frac{1}{p}} - R_n\bigg) \nonumber \\
&\leq \Pr\bigg(\Big(\frac{\sum_{i=1}^n X_i^p}{n}\Big)^{\frac{1}{p}} \geq \mu + 1/9 (\triangle/n)^{\frac{1}{p}}\bigg) + \Pr\bigg(\Big(\frac{2\sum_{i=1}^n Z_i}{n}\Big)^{\frac{1}{p}} \geq 4/9(\triangle/n)^{\frac{1}{p}}\bigg) \nonumber  (\text{ see $R_n \leq (4/9) (\triangle/n)^{\frac{1}{p}}$}) \\
&= \Pr\bigg(\Big(\frac{\sum_{i=1}^n X_i^p}{n}\Big)^{\frac{1}{p}} \geq \mu + \frac{1}{9} \frac{9}{4} (\frac{4}{9}\triangle_n/n)^{\frac{1}{p}}\bigg) + \Pr\bigg(\Big(\frac{2\sum_{i=1}^n Z_i}{n}\Big)^{\frac{1}{p}} \geq (\frac{(4/9)^{p}\triangle}{n})^{\frac{1}{p}}\bigg) \nonumber (\text{ definition of $\triangle$}) \\
&\leq \Pr\bigg(\Big(\frac{\sum_{i=1}^n X_i^p}{n}\Big)^{\frac{1}{p}} \geq \mu + \triangle_n/9n\bigg) + \Pr\bigg(\Big(\frac{\sum_{i=1}^n Z_i}{n}\Big) \geq 2\triangle_{n}/9n\bigg) \nonumber (\text{see } (\ref{13}) \text{ and } f(x) = a^x \text{ is decrease when $a < 1$})
\end{flalign}
The first term is bounded by $\delta/2$ according to (\ref{19}) and the second term is bounded by $\delta/2M$ according to Lemma \ref{lb:lemma6} (the condition of Lemma \ref{lb:lemma6} is satisfied because $a_n \leq (1/9)\triangle_n$). This finishes the proof of the first part (\ref{lb_lower}). The second part (\ref{lb_upper}) can be proved in an analogous manner. 
\end{proof}

\begin{manualtheorem} {5} \label{theorem5}
Fix an arbitrary $\delta \leq 0$ and fix $p \geq 1$, $M = \exp(H_{p,1})$ where $H_{p,1}$ is defined as in Lemma \ref{lb:lemma2} and let $\triangle_n = (\frac{9}{4})^{p-1} (9\sqrt{2n \log(2M/\delta)})$. Let $n_0$ be such that
\begin{flalign}
\sqrt{n_0} \leq \mathcal{O}(K(C^2 \log n_0 + N_0 (1/2))).
\end{flalign}
Then for any $n \geq n_0$, under the assumptions of Theorem 2, the following bounds hold true:
\begin{flalign}
\Pr(\overline{X}_{n}(p) \geq \E[ \overline{X}_{n}(p)] + (\triangle_n/n)^{\frac{1}{p}}) \leq \delta \\
\Pr(\overline{X}_{n}(p) \leq \E[ \overline{X}_{n}(p)] - (\triangle_n/n)^{\frac{1}{p}}) \leq \delta
\end{flalign}
\end{manualtheorem}
\begin{proof}


\noindent Let $X_t$ is the payoff sequence of the best arm. $Y_t$ is the payoff at time $t$. Both $X_t, Y_t$ lies in [0,1] interval, and\\ $\overline{X}_n(p) = \Big(\frac{\sum_{i=1}^n (1-Z_i) X_i^p + Z_i Y_i^p}{n}\Big)^{\frac{1}{p}}$
Apply Lemma \ref{lb:lemma6} and remember that $X^{\frac{1}{p}} - Y^{\frac{1}{p}} \leq (X-Y)^{\frac{1}{p}}$ we have:
\begin{flalign}
R_n &= \E\Bigg[\Big(\frac{\sum_{i = 1}^n X_i^p}{n}\Big)^{\frac{1}{p}}\Bigg] - \E\Bigg[\Big(\frac{\sum_{i=1}^n (1-Z_i) X_i^p + Z_i Y_i^p}{n}\Big)^{\frac{1}{p}}\Bigg]. \nonumber \\
&=\E\Bigg[\Big(\frac{\sum_{i = 1}^n X_i^p}{n}\Big)^{\frac{1}{p}} - \Big(\frac{\sum_{i=1}^n (1-Z_i) X_i^p + Z_i Y_i^p}{n}\Big)^{\frac{1}{p}}\Big].\nonumber \\
&\leq \E\Bigg[\Big(\frac{\sum_{i = 1}^n X_i^p - \sum_{i=1}^n (1-Z_i) X_i^p - Z_i Y_i^p}{n}\Big)^{\frac{1}{p}}\Bigg].\nonumber \\
&= \E\Bigg[\Big(\frac{\sum_{i=1}^n Z_i (X_i^p - Y_i^p)}{n}\Big)^{\frac{1}{p}}\Bigg].\nonumber \\
&\leq \E\Bigg[\Big(\sum_{i=1}^K \frac{T_i(n)}{n}\Big)^{\frac{1}{p}}\Bigg].\nonumber \\
&\leq \Bigg(\frac{\sum_{i=1}^K \E[T_i(n)]}{n}\Bigg)^{\frac{1}{p}}. \text{ see Jensen inequality}\nonumber \\ 
&= \Bigg((K-1) O\Bigg(\frac{K(C^2 \log n + N_0 (1/2))}{n}\Bigg)\Bigg)^{\frac{1}{p}}. \nonumber 
\end{flalign}
So that let $n_0$ be an index such that if $n \geq n_0$ then $a_n \leq \triangle_n/9$ and 
$R_n \leq 4/9(\triangle_n/n)^{\frac{1}{p}}$. Such an index exists since $\triangle_n = \mathcal{O}(\sqrt{n})$ and $a_n, R_n = \mathcal{O}((\log n/n)^{\frac{1}{p}})$. Hence, for $n \geq n_0$, the conditions of lemma \ref{lb:lemma6} are satisfied and the desired tail-inequalities hold for $\overline{X_n}(p)$.\\
\end{proof}
In the next theorem, we show that \alg~ can ensure the convergence of choosing the best arm at the root node.
\begin{manualtheorem} {6} (\textbf{Convergence of Failure Probability})
Under the assumptions of Theorem 2, it holds that
\begin{flalign}
\lim_{t\rightarrow \infty} \Pr(I_t \neq i^*) = 0
\end{flalign}
\end{manualtheorem}
\begin{proof}
We show that \alg~ can smoothly adapt to UCT's prove.
Let $i$ be the index of a suboptimal arm and let $p_{it} = \Pr(\overline{X}_{i,T_i(t)}(p) \geq \overline{X}^{*}_{T^*(t)}(p))$  from above. Clearly, $\Pr(I_t \neq i*) \leq \sum_{i\neq i*}p_{it}$. Hence, it suffices to show that $p_{it} \leq \epsilon/K$ holds for all suboptimal arms for t sufficiently large.\\
Clearly, if $\overline{X}_{i,T_i(t)}(p) \leq \mu_i + \triangle_i/2$ and $\overline{X}^*_{T^*(t)}(p) \geq \mu^{*} - \triangle_i/2$ then $\overline{X}_{i,T_i(t)}(p) < \overline{X}^{*}_{T^*(t)}(p)$. Hence,\\
\begin{flalign}
p_t \leq \Pr(\overline{X}_{i,T_i(t)}(p) \leq \mu_i + \triangle_i/2) + \Pr(\overline{X}^{*}_{T^*(t)}(p) \geq \mu^{*} - \triangle_i/2) \nonumber 
\end{flalign}
The first probability can be expected to be converging much slower since $T_i(t)$ converges slowly. Hence, we bound it first.\\
In fact, 
\begin{flalign}
\Pr(\overline{X}_{i,T_i(t)}(p) \leq \mu_i + \triangle_i/2) \leq \Pr(\overline{X}_{i,T_i(t)}(p) \leq \overline{\mu}_{i,T_i(t)} - |\delta_{i, T_i(t)}| + \triangle_i/2). \nonumber 
\end{flalign}
Without the loss of generality, we may assume that $|\delta_{i,T_i(t)}| \leq \triangle_i/4$. Therefore
\begin{flalign}
\Pr(\overline{X}_{i,T_i(t)}(p) \leq \mu_i + \triangle_i/2) \leq \Pr(\overline{X}_{i,T_i(t)}(p) \leq \overline{\mu}_{i,T_i(t)} + \triangle_i/4). \nonumber 
\end{flalign}
Now let a be an index such that if $t \geq a$ then $(t+1)\Pr(\overline{X}_{i,T_i(t)}(p) \leq \overline{\mu}_{i,T_i(t)} + \triangle_i/4) \leq \epsilon/(2K)$. Such an index exist by our assumptions on the concentration properties of the average payoffs. Then, for $t \geq a$
\begin{flalign}
\Pr(\overline{X}_{i,T_i(t)}(p) \leq \overline{\mu}_{i,T_i(t)} + \triangle_i/4) \leq \Pr(\overline{X}_{i,T_i(t)}(p) \leq \overline{\mu}_{i,T_i(t)} + \triangle_i/4, T_i(t) \geq a) + \Pr(T_i(t) \leq a) \nonumber 
\end{flalign}
Since the lower-bound on $T_i(t)$ grows to infinity as $t \rightarrow \infty$, the second term becomes zero when t is sufficiently large. The first term is bounded using the method of Lemma \ref{lb:lemma5}. By choosing $b = 2a$, we get
\begin{flalign}
\Pr(\overline{X}_{i,T_i(t)}(p) \leq \overline{\mu}_{i,T_i(t)} + \triangle_i/4, T_i(t) \geq a) \leq (a+1)\Pr(\overline{X}_{i,a}(p) \leq \overline{\mu}_{i,a} + \triangle_i/4, T_i(t) \geq a) + \Pr(T_i(t) \geq 2b) \leq \epsilon/(2K),  \nonumber 
\end{flalign}
where we have assumed that $t$ is large enough  so that $P(T_i(t) \geq 2b) = 0$.\\
Bounding $\Pr(\overline{X}^{*}_{T^*(t)}(p) \geq \mu^{*} - \triangle_i/2)$ by $\epsilon/(2K)$ can be done in an analogous manner. Collecting the bound yields that $p_{it} \leq \epsilon/K$ for $t$ sufficiently large which complete the prove.
\end{proof}
Now is our result to show the bias of expected payoff $\overline{X_n}(p)$
\begin{manualtheorem} {7}
Consider algorithm \alg \space running on a game tree of depth D, branching factor K with stochastic payoff
at the leaves. Assume that the payoffs lie in the interval [0,1]. Then the bias of the estimated expected
payoff, $\overline{X_n}$, is $\mathcal{O} (KD (\log (n)/n)^{\frac{1}{p}} + K^D (1/n)^{\frac{1}{p}})$. Further, the failure
probability at the root 
convergences to zero as the number of samples grows to infinity.
\end{manualtheorem}
\begin{proof}
The proof is done by induction on $D$.
When $D = 1$, \alg \space becomes UCB1 problem but using Power Mean backup instead of average mean and the convergence is guaranteed directly from Theorem \ref{theorem1}, Theorem~\ref{T:th_3} and Theorem $6$.

Now we assume that the result holds up to depth $D-1$ and consider the tree of Depth $D$.
Running \alg \space on root node is equivalence as UCB1 on non-stationary bandit settings. The error bound of
running \alg \space for the whole tree is the sum of payoff at root node with payoff starting from any node 
$i$ after the first action chosen from root node until the end. This payoff by induction at depth $(D-1)$ is
\begin{flalign}
\mathcal{O} (K(D-1) (\log (n)/n)^{\frac{1}{p}} + K^{D-1} (1/n)^{\frac{1}{p}}). \nonumber
\end{flalign}
According to the Theorem~\ref{T:th_3}, the payoff at the root node is 
\begin{flalign}
|\delta^*_n| + \mathcal{O} \Bigg( \frac{K(\log n + N_0)}{n} \Bigg)^{\frac{1}{p}}. \nonumber
\end{flalign}
The payoff of the whole tree with depth $D$:
\begin{flalign}
& |\delta^*_n| + \mathcal{O} \Bigg( \frac{K(\log n + N_0)}{n} \Bigg)^{\frac{1}{p}} \nonumber \\ 
&= \mathcal{O} (K(D-1) (\log (n)/n)^{\frac{1}{p}} + K^{D-1} (1/n)^{\frac{1}{p}}) \nonumber \\
&+ \mathcal{O} \Bigg( \frac{K(\log n + N_0)}{n} \Bigg)^{\frac{1}{p}} \nonumber \\
&\leq \mathcal{O} (K(D-1) (\log (n)/n)^{\frac{1}{p}} + K^{D-1} (1/n)^{\frac{1}{p}}) \nonumber \\
&+ \mathcal{O} \Bigg( K\left(\frac{\log n}{n}\right)^{\frac{1}{p}} + KN_0\left(\frac{1}{n}\right)^{\frac{1}{p}}\Bigg) \nonumber \\
&= \mathcal{O} (KD (\log (n)/n)^{\frac{1}{p}} + K^{D} (1/n)^{\frac{1}{p}})\nonumber
\end{flalign}
with $N_0 = O((K-1)K^{D-1})$, which completes our proof of the convergence of \alg. Since by our induction hypothesis this holds for all nodes at a distance of one node from the root, the proof is finished by observing that Theorem 3 and Theorem 5 do indeed ensure that the drift conditions are satisfied.
Interestingly, the proof guarantees the convergence for any finite value of $p$.
\end{proof}

\end{document}